\documentclass{article}

\usepackage[preprint]{neurips_2025}

\usepackage{amsmath,amsfonts,bm}

\def\eqref#1{equation~\ref{#1}}
\def\1{\bm{1}}

\DeclareMathAlphabet{\mathsfit}{\encodingdefault}{\sfdefault}{m}{sl}
\SetMathAlphabet{\mathsfit}{bold}{\encodingdefault}{\sfdefault}{bx}{n}

\newcommand{\KL}{D_{\mathrm{KL}}}

\usepackage{wrapfig}
\usepackage{amssymb}
\usepackage{amsmath}
\usepackage{bm}
\usepackage[utf8]{inputenc} % allow utf-8 input
\usepackage[T1]{fontenc}    % use 8-bit T1 fonts
\usepackage{hyperref}       % hyperlinks
\usepackage{url}            % simple URL typesetting
\usepackage{booktabs}       % professional-quality tables
\usepackage{amsfonts}       % blackboard math symbols
\usepackage{nicefrac}       % compact symbols for 1/2, etc.
\usepackage{microtype}      % microtypography
\usepackage{xcolor, colortbl}         % colors
\usepackage{subfig}
\usepackage{graphicx}
\usepackage{xspace}
\usepackage{ragged2e}
\usepackage{multirow}
\usepackage{wrapfig}

\definecolor{GrayBG}{gray}{0.95}
\definecolor{cluster}{RGB}{250, 215, 172}
\definecolor{entity}{RGB}{231,255,219}

\newcommand{\modelname}{\text{DSLFM-KGC}}

\title{Deep Sparse Latent Feature Models for \\ Knowledge Graph Completion}

\author{%
  \bf Haotian Li$^{1}$\thanks{Equal contribution.}~~Rui Zhang$^{1}$\footnotemark[1]~~Lingzhi Wang$^{1}$\footnotemark[1]~~Bin Yu$^{1}$~~Youwei Wang$^{1}$ \\
  \bf Yuliang Wei$^{1}$~~Kai Wang$^{1}$~~Richard Yi Da Xu$^{2}$\thanks{Co-corresponding authors.}~~Bailing Wang$^{1}$\footnotemark[2] \\
  $^1$ Harbin Institute of Technology, Harbin, China \\
  $^2$ Hong Kong Baptist University, Hongkong, China
}

\begin{document}

\maketitle

\begin{abstract}
Recent advances in knowledge graph completion (KGC) have emphasized text-based approaches to navigate the inherent complexities of large-scale knowledge graphs (KGs). While these methods have achieved notable progress, they frequently struggle to fully incorporate the global structural properties of the graph. Stochastic blockmodels (SBMs), especially the latent feature relational model (LFRM), offer robust probabilistic frameworks for identifying latent community structures and improving link prediction. This paper presents a novel probabilistic KGC framework utilizing sparse latent feature models, optimized via a deep variational autoencoder (VAE). Our proposed method dynamically integrates global clustering information with local textual features to effectively complete missing triples, while also providing enhanced interpretability of the underlying latent structures. Extensive experiments on four benchmark datasets with varying scales demonstrate the significant performance gains achieved by our method.
\end{abstract}

\section{Introduction}
The majority of real-world phenomena exhibit multifaceted characteristics. For instance, social networks are not merely a collection of isolated individuals but represent a complex web of interactions across various contexts. Knowledge graphs (KGs) organize information into triples $(h, r, t)$, where $h$ denotes the head entity, $t$ the tail entity, and $r$ the relationship, forming extensive semantic networks. However, real-world KGs like DBpedia \citep{auer2007dbpedia} and Wikidata \citep{vrandevcic2014wikidata} often suffer from incompleteness, missing key entities and relationships \citep{dong2014knowledge}. Knowledge graph completion (KGC) aims to infer this missing information, improving the utility and completeness of the graph. 

Early KGC research focused on knowledge graph embedding (KGE) techniques \citep{bordes2013translating, sun2018rotate, balavzevic2019tucker}, which aimed to learn low-dimensional embeddings for entities and relations, employing various scoring functions to evaluate triples. More recently, text-based methods leveraging pre-trained language models (PLMs) \citep{yao2019kg, wang2021structure, wang2022simkgc} have attained state-of-the-art results on large-scale datasets such as Wikidata5M \citep{wang2021kepler}. These approaches typically rely on transforming the head embedding $\mathbf{h}$ into the tail $\mathbf{t}$ through the relation $\mathbf{r}$. However, \textbf{textual descriptions often suffer from information scarcity due to length constraints. Consequently, global information, particularly the complex interrelationships among entity communities, remains underexploited}. As noted by \cite{stanley2019stochastic}, network topologies frequently exhibit dense intra-group connections and sparser inter-group connections. While graph neural networks (GNNs), particularly message passing neural networks (MPNNs), enhance graph representation learning by incorporating node neighborhood structures, they have been shown to underutilize neighborhood information in later studies \citep{zhang2022rethinking, li2023message}. Moreover, GNN-based KGC methods also fail to incorporate inherent clustering properties. Motivated by these observations, we propose a triple completion method that integrates global clustering information with local textual features.

Uncovering the latent structure of graph data is a central focus in statistical network analysis \citep{porter2009communities, latouche2010bayesian}. Stochastic blockmodels (SBMs) \citep{airoldi2008mixed, miller2009nonparametric, latouche2011overlapping} constitute a widely recognized class of probabilistic models that assign cluster memberships to graph nodes and are highly regarded in both academic and industrial contexts. A prominent variant is the latent feature relational model (LFRM), which permits nodes to belong to multiple groups and employs an Indian Buffet Process (IBP) prior on the node-community assignment matrix $Z$ to determine the number of latent communities. These models typically employ Markov chain Monte Carlo (MCMC) \citep{miller2009nonparametric} or variational inference \citep{zhu2016max} for latent variable inference. While DGLFRM \citep{mehta2019stochastic} enhances SBM inference by utilizing a deep sparse variational autoencoder (VAE) \citep{kingma2013auto}, \textbf{it is not specifically designed for KGC tasks and encounters difficulties when scaling to large graphs with hundreds of thousands or even millions of nodes}.

\textbf{Contributions}. We propose {\modelname}, a novel method for addressing the KGC challenge by utilizing latent community structures in KGs. Our main contributions are as follows: \textbf{i)} we design an end-to-end probabilistic model for KGC that integrates supplementary sparse clustering features into triple representation, implemented through a deep VAE \citep{kingma2013auto}; \textbf{ii)} {\modelname} exhibits robust performance and interpretability in completing missing triples by leveraging community-level interconnections in entities; and \textbf{iii)} the deep architecture facilitates scalable inference. Through extensive experiments on the UMLS, WN18RR, FB15k-237, and Wikidata5M datasets, \textbf{iv)} we showcase our model’s superior capability and scalability in managing KGC tasks and uncovering interpretable latent structures.

\section{Preliminaries}
\subsection{Latent feature relational model}
The SBMs \citep{holland1983stochastic, airoldi2008mixed, miller2009nonparametric} are fundamental approaches for analyzing relational data, where a graph with $N$ nodes is represented by a binary adjacency matrix $A \in \{0,1\}^{N \times N}$. In this matrix, $A_{i,j}=1$ indicates a link between node $i$ and node $j$. Each node $i$ in an SBM is associated with a one-hot latent variable $\mathbf{z}_i \in \{0,1\}^{K}$ to indicate its community membership, where $K$ is the number of node communities.

For scenarios where nodes belong to multiple communities, the OSBM \citep{latouche2011overlapping} adapts the latent indicator $\mathbf{z}_i$ into a multivariate Bernoulli vector consisting of $K$ independent Bernoulli variables, denoted as $\mathbf{z}_i \sim \mathcal{MB}(\mathbf{z} | \bm{\pi})$:
\begin{equation}
\mathcal{MB}(\mathbf{z} | \bm{\pi}) = \prod_{k=1}^K \text{Bernoulli}(z_k | \pi_k) = \prod_{k=1}^K \pi_k^{z_k} (1-\pi_k)^{1-z_k}
\end{equation}
where $\pi_k \in [0,1]$. The link probability between two nodes in OSBM is defined as a bilinear function of their indicator vectors:
\begin{equation}
p(A_{i,j}=1 | \mathbf{z}_i, \mathbf{z}_j, W) = \sigma(\mathbf{z}_i^{\top} W \mathbf{z}_j)
\end{equation}
Here, $W$ is a real-valued $K \times K$ matrix, with $w_{kl}$ influencing the link likelihood between communities $k$ and $l$, and $\sigma(\cdot)$ is the sigmoid function.

Expanding on OSBM, the LFRM integrates the IBP prior \citep{miller2009nonparametric} on the binary node-community matrix $Z = [\mathbf{z}_1, \ldots, \mathbf{z}_N]^{\top}$, enabling dynamic learning of the number of communities. Traditional inference methods used in SBMs, such as MCMC or variational inference, often struggle to scale in large networks. To overcome this, DGLFRM \citep{mehta2019stochastic} uses a VAE \citep{kingma2013auto}, employing a graph convolutional network (GCN) \citep{kipf2016semi} to encode the variational distribution $q(Z)$ and a non-linear multilayer perceptron (MLP) to model the link probability $p(A_{i,j} | \mathbf{z}_i, \mathbf{z}_j, W)$. Despite its advances, DGLFRM encounters difficulties when applied to large-scale heterogeneous KGs, which feature entities and relations of diverse types.

\subsection{Knowledge graph completion}
A KG is commonly defined as $\mathcal{G} = (\mathcal{E}, \mathcal{R}, \mathcal{T})$, where $\mathcal{E}$ is the set of entities, and $\mathcal{R}$ is the set of relations. The set $\mathcal{T} = \{ (h,r,t) | h,t \in \mathcal{E}, r \in \mathcal{R} \}$ contains factual triples, each representing a directed labeled edge $h \mathop{\rightarrow}\limits^{r} t$ in the KG. Moreover, modern KGs often include meta-information $\mathcal{M}$, such as natural language descriptions \citep{yao2019kg, wang2022simkgc} or multi-modal data \citep{zhang2024native}. For any entity $e \in \mathcal{E}$ and any relation $r \in \mathcal{R}$, $\mathcal{M}(e)$ and $\mathcal{M}(r)$ denote the corresponding meta-information.

For a given query $(h,r,?)$, the task of KGC entails identifying the missing tail entity by retrieving the most plausible candidate $\hat{t}$ from the entity set $\mathcal{E}$, such that $(h,r,\hat{t})$ is valid. From the KGC perspective, we model a KG as comprising a query set $\mathcal{Q} = \{(h,r) | h \in \mathcal{E}, r \in \mathcal{R}\}$, a candidate answer (entity) set $\mathcal{E}$, and a mapping $\mathcal{A}: \mathcal{Q} \times \mathcal{E} \rightarrow \{0,1\}$ that determines whether a query has a valid answer in the KG $\mathcal{G}$. This mapping is represented as a binary matrix $A \in \{0,1\}^{|\mathcal{Q}| \times |\mathcal{E}|}$, analogous to an adjacency matrix, where $A_{hr,t}=1$ if the triple $(h,r,t) \in \mathcal{T}$, and $A_{hr,t}=0$ otherwise.

\section{Methodology}
This section presents the framework of {\modelname}. We begin by describing the probabilistic framework for KGs, emphasizing its application to KGC. Following this, we elaborate the VAE architecture employed for inference, detailing the design and implementation of both the encoder and decoder.

\subsection{Generative model}
We assume that triples within a KG are conditionally independent, given their latent communities. The generative process of a KG unfolds as follows:

For each query $(h,r) \in \mathcal{Q}$ and each answer $t \in \mathcal{E}$, draw the membership indicator vectors:
\begin{equation}
\mathbf{z}_{hr} \sim \mathcal{MB}(\mathbf{z} | \bm{\pi}_{hr}), \
\mathbf{z}_{t} \sim \mathcal{MB}(\mathbf{z} | \bm{\pi}_t)
\end{equation}

Next, draw the latent feature vectors:
\begin{equation}
\mathbf{w}_{hr} \sim \mathcal{N}(\mathbf{w} | \mathbf{0}, \sigma^2 \mathbf{I}), \ \mathbf{w}_{t} \sim \mathcal{N}(\mathbf{w} | \mathbf{0}, \sigma^2 \mathbf{I})
\end{equation}

finally, draw the triple:
\begin{equation}
A_{hr,t} \sim p(A_{hr,t} | \mathbf{z}_{hr}, \mathbf{z}_{t}, \mathbf{w}_{hr}, \mathbf{w}_{t})
\end{equation}

Here, $\mathbf{z}_{hr} \in \{0,1\}^{K_{1}}, \mathbf{z}_{t} \in \{0,1\}^{K_{2}}$ are binary vectors with elements equal to one indicating their respective community memberships, and $\mathbf{w}_{hr} \in \mathbb{R}^{K_{1}}, \mathbf{w}_{t} \in \mathbb{R}^{K_{2}}$ represent the strength of their community memberships, \emph{i.e}., latent features. Typically, query clusters outnumber entity clusters due to the diversity of entity-relation pairs.

The distribution $p(A_{hr,t} | \mathbf{z}_{hr}, \mathbf{z}_{t}, \mathbf{w}_{hr}, \mathbf{w}_{t})$ is modeled as a Bernoulli distribution, with the probability $p(A_{hr,t}=1 | \mathbf{z}_{hr}, \mathbf{z}_{t}, \mathbf{w}_{hr}, \mathbf{w}_{t})$ signifying that the answer aligns with the query, thus confirming the existence of the triple $(h,r,t)$ in the KG:
\begin{gather}
\mathbf{f}_{hr} = \mathbf{w}_\text{hr} \odot \mathbf{z}_{hr}, \ \mathbf{f}_{t} = \mathbf{w}_\text{t} \odot \mathbf{z}_{t} \label{eq:hadamard_product} \\
p(A_{hr,t}=1 | \mathbf{z}_{hr}, \mathbf{z}_{t}, \mathbf{w}_{hr}, \mathbf{w}_{t}) = \sigma\left( \mathbf{f}_{hr}^{\top} \mathbf{f}_{t} \right) \label{eq:p_x}
\end{gather}
where $\odot$ is the Hadamard product.

Let $Z_\text{qry}$ and $Z_\text{ans}$ denote the membership indicator matrices for queries and answers, respectively, and let $W_\text{qry}$ and $W_\text{ans}$ denote the latent feature matrices. Then, $F_\text{qry} = Z_\text{qry} \odot W_\text{qry}$ and $F_\text{ans} = Z_\text{ans} \odot W_\text{ans}$ constitute a sparse latent feature model \citep{ghahramani2005infinite, d2004direct, jolliffe2003modified}. We use the Indian Buffet Process (IBP) \citep{griffiths2011indian} prior on the indicator matrices to facilitate the learning of the number of communities, thereby establishing an infinite latent feature model \citep{ghahramani2005infinite}.
\begin{equation}
Z_\text{qry} \sim \mathcal{IBP}(\alpha_\text{qry}),\ Z_\text{ans} \sim \mathcal{IBP}(\alpha_\text{ans})
\end{equation}

\begin{figure*}[t]
  \centering
  \includegraphics[width=1\linewidth]{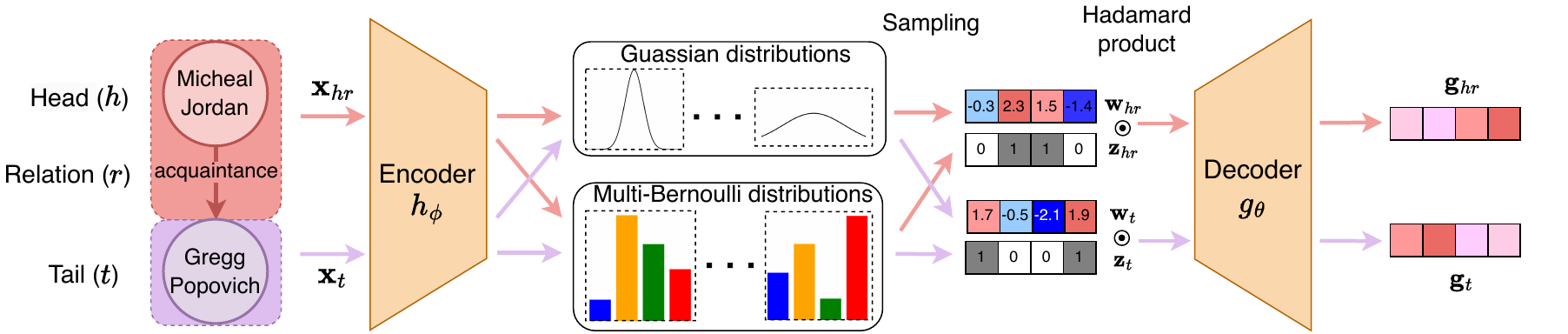}
  \caption{An overview of our {\modelname} framework during inference. Initially, the encoder network $h_\phi$ encodes the textual information of a triple ($\mathbf{x}_{hr}$ and $\mathbf{x}_{t}$) into posterior distributions, as defined in Equations \ref{eq:bert_encoding} and \ref{eq:mlp}. Latent variables (e.g., $\mathbf{z}_{hr}$ and $\mathbf{w}_{hr}$) are then sampled using reparameterization tricks (see Appendix \ref{app:reparam}), after which the decoder $g_\theta$ generates representations for the query and answer ($\mathbf{g}_{hr}$ and $\mathbf{g}_{t}$).}
  \label{fig:model_overview}
  \vspace*{-3mm}
\end{figure*}

\subsection{VAE encoder}
We adopt the stick-breaking construction of the IBP \citep{teh2007stick} to model $\mathbf{z}_{hr}$:
\begin{align}
v_{{hr}, k} & \sim \text{Beta}(\alpha_\text{qry}, 1),\ k=1, \ldots, K_1 \notag \\
\pi_{{hr}, k} & = \prod_{j=1}^k v_{{hr}, j},\ z_{hr,k} \sim \text{Bernoulli}(\pi_{{hr}, k}) \label{eq:pi_z}
\end{align}
The sampling of $\mathbf{z}_{t}$ can be achieved similarly. By employing the stick-breaking approach, the effective number of communities engaged can be learned by setting a sufficiently large truncation level $K = K_1 = K_2$ in our model.

Let $\mathcal{H} = \{V_\text{qry}, Z_\text{qry}, W_\text{qry}, V_\text{ans}, Z_\text{ans}, W_\text{ans}\}$ denote the set of latent variables and $\mathcal{O} = \{\mathcal{Q}, \mathcal{E}, A\}$ the set of observations. We utilize an encoder network to approximate the true posterior $p(\mathcal{H}|\mathcal{O})$ with a variational distribution $q_\phi(\mathcal{H})$ parameterized by $\phi$, which is factorized following the mean-field approximation:
\begin{equation}
q_\phi(\mathcal{H}) = q_\phi(\mathcal{H}_\text{qry}) q_\phi(\mathcal{H}_\text{ans})
\end{equation}
where $\mathcal{H}_\text{qry}$ and $\mathcal{H}_\text{ans}$ denote the query and answer latent variables, respectively. $q_\phi(\mathcal{H}_\text{qry})$ and $q_\phi(\mathcal{H}_\text{ans})$ are defined as follows:

\vspace*{-3mm}
\begin{equation}
\begin{aligned}
q_\phi(\mathcal{H}_\text{qry}) & = \prod_{(h,r) \in \mathcal{Q}} \prod_{k=1}^{K} q_\phi(v_{hr,k}) q_\phi(z_{hr,k}) q_\phi(w_{hr,k}) \\
q_\phi(\mathcal{H}_\text{ans}) & =  \prod_{t \in \mathcal{E}} \prod_{k=1}^{K} q_\phi(v_{t,k}) q_\phi(z_{t,k}) q_\phi(w_{t,k})
\end{aligned}
\end{equation}

The distributions involved are defined as follows:
\begin{align}
q_\phi(v_{hr,k}) & \triangleq \text{Beta}(c_{hr,k}, d_{hr,k}) \label{eq:q_vk} \\
q_\phi(z_{hr,k}) & \triangleq q_\phi(z_{hr,k} | \mathcal{Q}) = \text{Bernoulli}(\pi_{hr,k}(\mathcal{Q})) \label{eq:q_zk} \\
q_\phi(w_{hr,k}) & \triangleq q_\phi(w_{hr,k} | \mathcal{Q}) = \mathcal{N}(\mu_{hr,k}(\mathcal{Q}), \sigma^2_{hr,k}(\mathcal{Q})) \label{eq:q_wk}
\end{align}
where $\pi_{hr,k}, \mu_{hr,k}$ and $\sigma^2_{hr,k}$ are outputs of the encoder network $h_\phi$, \emph{i.e}., $\{ \bm{\pi}_{hr}, \bm{\mu}_{hr}, \bm{\sigma}^2_{hr}\}_{(h,r) \in \mathcal{Q}} = h_\phi(\mathcal{Q})$ with $\mathcal{Q}$ as the input. In experiment, we observed that treating $\mathbf{c}_{hr}$ and $\mathbf{d}_{hr}$ as part of the encoder parameters (instead of encoding them from the posterior) helps mitigate over-parameterization. We define $q_\phi(v_{t,k})$, $q_\phi(z_{t,k})$ and $q_\phi(w_{t,k})$ in a similar vein, with $\mathcal{Q}$ replaced by $\mathcal{E}$.

Following recent progress in text-based approaches for the KGC task \citep{yao2019kg,wang2022simkgc}, we employ the strategy that individually encodes the textual descriptions of queries and answers using two BERT \citep{devlin-etal-2019-bert} encoders, sharing pre-trained weights, and applying mean pooling:
\begin{equation}
\label{eq:bert_encoding}
\mathbf{e}_{hr} = \text{Pool}(\text{BERT}_\text{qry}(\mathbf{x}_{hr})), ~~
\mathbf{e}_{t} = \text{Pool}(\text{BERT}_\text{ans}(\mathbf{x}_t))
\end{equation}

Here, $\mathbf{x}_{hr}$ and $\mathbf{x}_t$ represent the textual descriptions of the query and the answer after tokenization, respectively. Subsequently, a multi-layer perceptron (MLP) is leveraged to project the textual encodings into the latent space:
\begin{equation}
\label{eq:mlp}
\{\pi_{hr,k}, \mu_{hr,k}, \sigma_{hr,k}\}_{k=1}^{K} = \text{MLP}(\mathbf{e}_{hr}), ~~
\{\pi_{t,k}, \mu_{t,k}, \sigma_{t,k}\}_{k=1}^{K} = \text{MLP}(\mathbf{e}_{t})
\end{equation}

It is important to note that the flexibility of our model allows for the use of various types of encoders, such as a multi-modal one, to further enhance expressiveness. We plan to explore these possibilities in future research.

We denote the overall encoder network with parameters $\phi$ as $h_\phi$. Integrating textual inputs not only enhances our model's performance, but also provides deeper insights into the latent structure. This allows for the exploration of the mined communities through the descriptions of their constituent entities, the benefits of which will be demonstrated in the experiment section.

\vspace*{-2.5mm}
\subsection{VAE decoder}
\vspace*{-1.5mm}
We model the probability distribution $p_\theta$ through a decoder network $g_\theta$, parameterized by $\theta$. Given the latent variables $\mathbf{z}_{hr}, \mathbf{z}_{t}, \mathbf{w}_{hr}$ and $\mathbf{w}_{t}$, the decoder network generates a link $A_{hr, t} \sim p_\theta(A_{hr, t} | \mathbf{z}_{hr}, \mathbf{z}_{t}, \mathbf{w}_{hr}, \mathbf{w}_{t})$. We first computes the Hadamard product to obtain $\mathbf{f}_{hr}$ and $\mathbf{f}_{t}$, as outlined in Equation \ref{eq:hadamard_product}. An MLP with non-linear activations is subsequently employed to transform $\mathbf{f}_{hr}, \mathbf{f}_{t}$ into $\mathbf{g}_{hr}, \mathbf{g}_{t}$, respectively.
\begin{align}\label{eq:mlp_decoding}
\mathbf{g}_{hr} = \text{MLP}(\mathbf{f}_{hr}), ~~ \mathbf{g}_{t} = \text{MLP}(\mathbf{f}_{t})
\end{align}
The inner product of these transformed vectors is then computed to represent the confidence level that the triple $(h,r,t)$ exists in the KG. The use of an MLP, as opposed to relying solely on a single Hadamard product, enables more expressive representations and improves overall performance.

The architecture of our model is depicted in Figure \ref{fig:model_overview}.

\subsection{Inference}
We jointly update the encoder $h_\phi$ and the decoder $g_\theta$ by minimizing the negative of the evidence lower bound (ELBO):

\vspace*{-5mm}
\begin{align}\label{eq:elbo}
\mathcal{L}
& = \KL [q_\phi(\mathcal{H}) || p_\theta(\mathcal{H})] - \mathbb{E}_q \left[ \log p_\theta(\mathcal{O} |\mathcal{H}) \right] \notag \\
& = \KL[q_\phi(\mathcal{H}_\text{qry}) || p_\theta(\mathcal{H}_\text{qry})] + \KL[q_\phi(\mathcal{H}_\text{ans}) || p_\theta(\mathcal{H}_\text{ans})] \notag  \\
& ~ - \mathbb{E}_q \left[ \log p_\theta(\mathcal{Q} |\mathcal{H}_\text{qry}) \right] - \mathbb{E}_q \left[ \log p_\theta(\mathcal{E} |\mathcal{H}_\text{ans}) \right] - \mathbb{E}_q \left[ \log p_\theta(A |\mathcal{H}) \right]
\end{align}

where $\KL[q(\cdot) || p(\cdot)]$ is the KL divergence of the distributions $q(\cdot)$ and $p(\cdot)$.

To further enhance KGC performance, we express the triple completion term $\log p_\theta(A |\mathcal{H})$ as a contrastive loss for its renowned capacity to learn expressive representations, which aims to maximize the mutual information between the inputs and the outputs \citep{ben2023reverse, hjelm2018learning, gutmann2012noise}. Specifically, given the set of latent variables for a specific query and answer, we utilize the supervised contrastive loss \citep{li2023kermit, khosla2020supervised}:

\vspace*{-2mm}
\begin{equation}
\log p_\theta(A_{hr,t} | \mathbf{z}_{hr}, \mathbf{z}_{t}, \mathbf{w}_{hr}, \mathbf{w}_{t}) = \frac{1}{|\mathcal{N}^+|} \sum_{t \in \mathcal{N}^+} \log \frac{e^{S(\mathbf{g}_{hr}, \mathbf{g}_{t})}}{e^{S(\mathbf{g}_{hr}, \mathbf{g}_{t})} + \sum_{t' \in \mathcal{N}^-}{e^{S(\mathbf{g}_{hr}, \mathbf{g}_{t})}}}
\end{equation}

where $\mathcal{N}^+$ represents the set of positive entities of the query $(h,r,?)$, and $\mathcal{N}^-$ denotes the set of negative samples, encompassing all other entities within the same batch \citep{pmlr-v119-chen20j}. $S(\mathbf{g}_{hr}, \mathbf{g}_t) = (\cos(\mathbf{g}_{hr}, \mathbf{g}_t) - \gamma) / \tau$ is the cosine similarity score function with additive margin $\gamma$ and temperature $\tau$, following the work of \citep{kumar2018mises}.

We then optimize the objective using stochastic gradient variational Bayes (SGVB) and mini-batch gradient descent \citep{kingma2013auto}. Given a batch of triples $B \subset \mathcal{G}$, and let the decoded representations $\mathbf{g}_{hr}, \mathbf{g}_{t} \in \mathbb{R}^D$, the computation of $\mathcal{L}$ requires time $\mathcal{O}(|B| \cdot (C_\text{KL}+C_\text{Recon}+C_\text{Comp}))$ and space $\mathcal{O}(|B| \cdot D + |B| \cdot K)$, where $C_\text{KL}$, $C_\text{Recon}$ and $C_\text{Comp}$ denotes the complexity of evaluating the KL divergence, reconstruction and triple completion terms in the ELBO, respectively. Please refer to Appendix \ref{app:math} for additional proofs and computation details.

\section{Experiment}
\begin{table*}[t]
  \caption{
    KGC results for the WN18RR, FB15k-237 and UMLS datasets.
  }
  \centering
  \resizebox{\textwidth}{!}{\begin{tabular}{c|cccc|cccc|cccc}
    \toprule[1pt]
    \midrule
    \multicolumn{1}{c|}{\multirow{2}{*}{Method}} & \multicolumn{4}{c|}{WN18RR} & \multicolumn{4}{c}{FB15k-237} & \multicolumn{4}{c}{UMLS} \\
    & MRR & Hit@1 & Hit@3 & Hit@10 & MRR & Hit@1 & Hit@3 & Hit@10 & MRR & Hit@1 & Hit@3 & Hit@10 \\
    \midrule
    \multicolumn{13}{c}{\textit{Embedding-based Methods}} \\
    \midrule
    TransE \cite{bordes2013translating} & 24.3 & 4.3 & 44.1 & 53.2 & 27.9 & 19.8 & 37.6 & 44.1 & 61.5 & 39.1 & - & 98.9 \\
    DistMult \cite{yang2014embedding} & 44.4 & 41.2 & 47.0 & 50.4 & 28.1 & 19.9 & 30.1 & 44.6 & 39.1 & 25.6 & 44.5 & 66.9 \\
    RotatE \cite{sun2018rotate} & \underline{47.6} & 42.8 & \underline{49.2} & \underline{57.1} & 33.8 & 24.1 & 37.5 & 53.3 & 74.4 & 63.6 & 82.2 & 93.9 \\
    ConvE \cite{dettmers2018convolutional} & 43.0 & 40.0 & 44.0 & 52.0 & 32.5 & 23.7 & 35.6 & 50.1 & 83.6 & 76.4 & \underline{96.0} & 99.0 \\
    TuckER \cite{balavzevic2019tucker} & 47.0 & \underline{44.3} & 48.2 & 52.6 & \underline{35.8} & \underline{26.6} & \underline{39.4} & \underline{54.4} & 73.2 & 62.5 & 81.2 & 90.0 \\
    HittER \cite{chen-etal-2021-hitter} & 50.3 & 46.2 & 51.6 & 58.4 & \underline{37.3} & \underline{27.9} & \underline{40.9} & \textbf{55.8} & - & - & - & - \\
    KRACL \cite{tan2023kracl} & \underline{52.7} & \underline{48.2} & \underline{54.7} & \underline{61.3} & 36.0 & 26.6 & 39.5 & 54.8 & \underline{90.4} & \underline{83.1} & - & \underline{99.5} \\
    \midrule
    \multicolumn{13}{c}{\textit{GNN-based Methods}} \\
    \midrule
    RGCN \cite{schlichtkrull2018modeling} & 12.3 & 8.0 & 13.7 & 20.7 & 16.4 & 10.0 & 18.1 & 30.0 & \underline{48.1} & \underline{31.8} & - & \underline{83.5} \\
    CompGCN \cite{vashishth2019composition} & \underline{47.2} & \underline{43.7} & \underline{48.5} & \underline{54.0} & \underline{35.5} & \underline{26.4} & \underline{39.0} & \underline{53.6} & - & - & - & - \\
    KGGAT \cite{nathani2019learning} & 46.4 & 42.6 & 47.9 & 53.9 & 35.0 & 26.0 & 38.5 & 53.1 & - & - & - & - \\
    \midrule
    \multicolumn{13}{c}{\textit{Text-based Methods}} \\
    \midrule
    KG-BERT \cite{yao2019kg} & 21.6 & 4.1 & 30.2 & 52.4 & - & - & - & 42.0 & - & - & - & 99.0 \\
    MTL-KGC \citep{kim-etal-2020-multi} & 33.1 & 20.3 & 38.3 & 59.7 & 26.7 & 17.2 & 29.8 & 45.8 & - & - & - & - \\
    StAR \citep{wang2021structure} & 40.1 & 24.3 & 49.1 & 70.9 & 29.6 & 20.5 & 32.2 & 48.2 & - & - & - & \underline{99.1} \\
    SimKGC \cite{wang2022simkgc} & 66.6 & 58.7 & 71.7 & 80.0 & 33.6 & 24.9 & 36.2 & 51.1 & \underline{79.4} & \underline{70.4} & \underline{86.5} & 94.4 \\
    KG-S2S \citep{chen2022knowledge} & 57.4 & 53.1 & 59.5 & 66.1 & 33.6 & \underline{25.7} & \underline{37.3} & 49.8 & - & - & - & - \\
    GHN \citep{qiao-etal-2023-improving} & \underline{67.8} & \underline{59.6} & \underline{71.9} & \underline{82.1} & \underline{33.9} & 25.1 & 36.4 & \underline{51.8} & - & - & - & - \\
    \midrule
    \multicolumn{13}{c}{\textit{LLM-based Methods}} \\
    \midrule
    CP-KGC \cite{yang2024enhancing} & \underline{67.3} & \underline{59.9} & \underline{72.1} & \underline{80.4} & 33.8 & 25.1 & 36.5 & 51.6 & - & - & - & - \\
    KICGPT \cite{wei2024kicgpt} & 54.9 & 47.4 & 58.5 & 64.1 & \textbf{41.2} & \textbf{32.7} & \textbf{44.8} & \underline{55.4} & - & - & - & - \\
    \midrule
    \rowcolor{GrayBG} {\modelname} (ours) & \textbf{70.4} & \textbf{63.1} & \textbf{74.8} & \textbf{84.2} & \underline{35.5} & \underline{26.4} & \underline{38.9} & \underline{53.7} & \textbf{92.4} & \textbf{86.4} & \textbf{98.7} & \textbf{99.6} \\
    \midrule
    \bottomrule[1pt]
  \end{tabular}}
  \label{tab:kgc_results_wn_fb}
  \vspace*{-5mm}
\end{table*}

\subsection{Experiment settings}
\textbf{Datasets}. To evaluate our method for filling in missing triples in KGs, we selected benchmark datasets ranging from small-sized (about 6k triples) to large-scale (around 20 million triples) for the KGC task. These include UMLS \cite{kok2007statistical}, WN18RR, FB15k-237 \citep{toutanova2015representing}, and Wikidata5M-Ind \citep{wang2021kepler}.

\textbf{Evaluation metrics}. In our approach, for each query $(h,r,?)$, a score is calculated for each entity and the rank of the correct answer is determined. We report the Mean Reciprocal Rank (MRR) and Hit@$k$ metrics under the filtered protocol \citep{bordes2013translating}. For each triple $(h,r,t)$, we construct a forward query $(h,r,?)$ with $t$ as the answer, along with a backward query $(?,r^{-1},t)$ for data augmentation. Here, $r^{-1}$ denotes the inverse of the relation $r$, as sourced from \cite{li2023kermit}. The averaged results of the forward and backward metrics are reported in our experimental evaluations.

\textbf{Baselines}. We conduct comprehensive experiments to evaluate the performance of our model against a variety of KGC models, encompassing embedding-based, GNN-based, text-based and LLM-based KGC approaches.

\textbf{Implementation details}. To ensure a fair comparison with existing approaches, we maintain the same primary hyperparameters. Specifically, the BERT encoders are initialized with pre-trained weights from "bert-base-uncased". We use a batch size of 1024 with 4 Quadro RTX 8000 GPUs, although a larger batch size is reasonably expected to provide better performance under the contrastive framework. The maximum number of communities $K$ is consistently set to 128 for all datasets. In the case of the WN18RR, FB15k-237 and UMLS datasets, we utilize in-batch negative sampling, whereas for the Wikidata5M-Ind dataset, we adopt an additional self-negative sampling strategy to ensure fair comparison with SimKGC \citep{wang2022simkgc}.

Detailed information regarding the experimental setup can be found in Appendix \ref{app:experiment_settings}.

\vspace*{-2mm}
\subsection{Main results}\label{sec:main_results}
\begin{wrapfigure}{r}{0.5\textwidth}
    \vspace*{-5mm}
    \centering
    \caption{KGC results for the Wikidata5M-Ind dataset.}
    \vspace*{-2mm}
    \label{tab:kgc_results_wiki}
    \resizebox{70mm}{!}{
      \begin{tabular}{c|cccc}
        \toprule[1pt]
        \midrule
        Method & MRR & Hit@1 & Hit@3 & Hit@10 \\
        \midrule
        DKPL \citep{xie2016representation} & 23.1 & 5.9 & 32.0 & 54.6 \\
        KEPLER \citep{wang2021kepler} & 40.2 & 22.2 & 51.4 & 73.0 \\
        BLP-ComplEx \cite{daza2021inductive} & 48.9 & 26,2 & 66.4 & 87.7 \\
        BLP-SimplE \cite{daza2021inductive} & 49.3 & 28.9 & 63.9 & 86.6 \\
        SimKGC \cite{wang2022simkgc} & \underline{71.3} & \underline{60.7} & \underline{78.7} & \underline{91.3} \\
        \midrule
        \rowcolor{GrayBG} DSLFM-KGC & \textbf{76.3} & \textbf{67.2} & \textbf{82.7} & \textbf{93.6} \\
        \midrule
        \bottomrule[1pt]
      \end{tabular}
    }
    \vspace*{-3.5mm}
\end{wrapfigure}

Owing to the stochastic nature of our model, we conduct five independent experiments with distinct random seeds and report the average metrics. Table~\ref{tab:kgc_results_wn_fb} presents the results for the WN18RR and FB15k-237 datasets, while Table~\ref{tab:kgc_results_wiki} presents the results for the Wikidata5M-Ind dataset. Hit@k is expressed as a percentage. The highest performance for each metric on each dataset is indicated in bold, and the top-performing metrics across categories are underlined.

The most substantial improvement is observed on the Wikidata5M-Ind dataset, where our model exhibits a 5.0\% increase in MRR (from 71.3\% to 76.3\%) and a 6.5\% increase in Hit@1 (from 60.7\% to 67.2\%) compared to SimKGC. Similar improvements are observed on the WN18RR dataset, where {\modelname} surpasses the second-best model (GHN) across all metrics, with enhancements ranging from 1.9\% to 3.5\% in MRR and Hit@k, demonstrating its strong predictive capability. On the FB15k-237 dataset, while our model falls short of embedding-based models and KICGPT \cite{wei2024kicgpt}, it still outperforms GNN-based and text-based methods, narrowing the gap between text-based and embedding-based approaches by approximately 2-3 percentage points. \textbf{The results demonstrate both the effectiveness and the scalability of our model.}

To elucidate the results obtained from the WN18RR and FB15k-237 datasets, we perform a detailed analysis of the underlying KGs. \textbf{i)} First, we assess the topological structure of each KG by calculating the average degree $M/N$, where $M$ and $N$ represent the number of edges and nodes, respectively. The FB15k-237 dataset exhibits a denser structure, with an average degree of 21.3, compared to 2.27 for WN18RR. \textbf{ii)} Second, we measure the modularity of the KGs and present the results in Appendix \ref{app:analysis}. The significantly lower modularity score for FB15k-237 (0.074) quantitatively confirms our hypothesis that \textbf{FB15k-237 exhibits much less pronounced clustering patterns}. \textbf{iii)} Finally, we conduct in-depth ablation and analysis studies to further investigate the challenges our model encounters when capturing latent community structures from the FB15k-237 dataset, as discussed in subsequent sections.

\vspace*{-3mm}
\subsection{Ablation results}
We conduct a series of ablation experiments to investigate the impact of key model components on KGC performance.

\begin{wrapfigure}{r}{0.5\textwidth}
  \vspace*{-5mm}
  \centerline{
      \includegraphics[width=\linewidth]{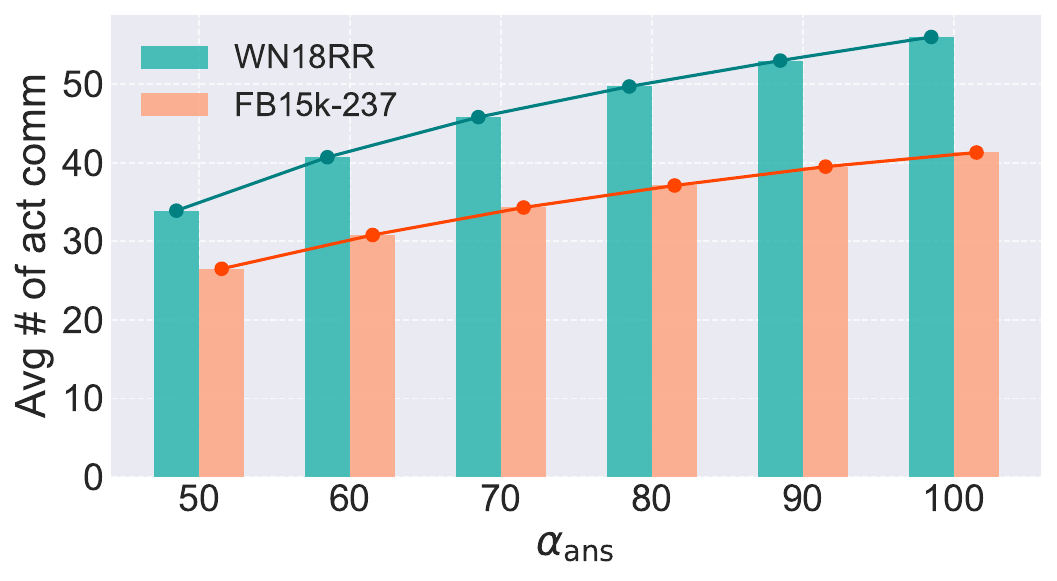}
  }
  \caption{Average number of activated communities learned on the WN18RR and FB15k-237 datasets.
  }
  \label{fig:ablation_avg_comm_number}
  \vspace*{-4mm}
\end{wrapfigure}
\textbf{Robustness \emph{w.r.t}. hyperparameter settings}. We conduct KGC experiments with $\alpha_\text{qry}$ and $\alpha_\text{ans}$ selected from the grid $\{80, 90, 100\} \times \{10, 20, \ldots, 100\}$, while maintaining all other hyperparameters fixed. Table \ref{tab:ablation_alpha} reports the mean and standard deviation of these 30 results for each dataset. The minimal variability observed across different $\alpha_\text{qry}$ and $\alpha_\text{ans}$ values, as seen in Table \ref{tab:ablation_alpha}, underscores the \textbf{robustness of our model to varying prior settings, confirming that our model maintains a favorable balance between completing the KGC task and determining the number of clusters}. Further analysis of the model's robustness to other hyperparameters is provided in Appendix \ref{app:ablation}.

As discussed in Section \ref{sec:main_results}, the denser topology of the FB15k-237 dataset presents a greater challenge for capturing community structures. To gain further insight, we compute the average number of activated communities (the number of non-zero entries in $Z_\text{ans}$ divided by the number of entities $|\mathcal{E}|$) and present the trend across varying $\alpha_\text{ans}$ values in Figure \ref{fig:ablation_avg_comm_number}. Clearly, for identical $\alpha_\text{ans}$ values, FB15k-237 exhibits significantly fewer latent communities than WN18RR, with the disparity increasing as $\alpha_\text{ans}$ rises. This indicates the greater density and less pronounced clustering structure of the FB15k-237 dataset.

\begin{table}[t]
    % \vspace*{-4mm}
    \centering
    \caption{Performance of {\modelname} on the WN18RR, FB15k-237 and UMLS datasets w/ different latent structures.}
    \vspace*{1mm}
    \label{tab:ablation_latent_structure}
    \resizebox{\linewidth}{!}{
      \begin{tabular}{c|ccc|ccc|ccc}
        \toprule[1pt]
        \midrule
        \multirow{2}{*}{Method} & \multicolumn{3}{c|}{WN18RR} & \multicolumn{3}{c}{FB15k-237} & \multicolumn{3}{c}{UMLS} \\
        & Hit@1 & Hit@10 & Epochs & Hit@1 & Hit@10 & Epochs & Hit@1 & Hit@10 & Epochs \\
        \midrule
        \rowcolor{GrayBG} Ours & \textbf{63.1} & \textbf{84.2} & 65 & \textbf{26.4} & \textbf{53.7} & 15 & \textbf{86.4} & \textbf{99.6} & 25 \\
        VAE & \underline{59.1} & \underline{81.0} & 55 & \underline{25.4} & 51.5 & 10 & \underline{71.5} & 89.0 & 23 \\
        AE & 57.9 & 80.0 & 50 & 25.2 & \underline{52.0} & 10 & 69.6 & 92.7 & 19 \\
        \midrule
        SimKGC & 58.7 & 80.8 & 50 & 24.9 & 51.1 & 10 & 70.4 & \underline{94.4} & 20 \\
        \midrule
        \bottomrule[1pt]
      \end{tabular}
    }
    \vspace*{-6mm}
\end{table}

\begin{figure}[t]
  \centering
  \includegraphics[width=1\linewidth]{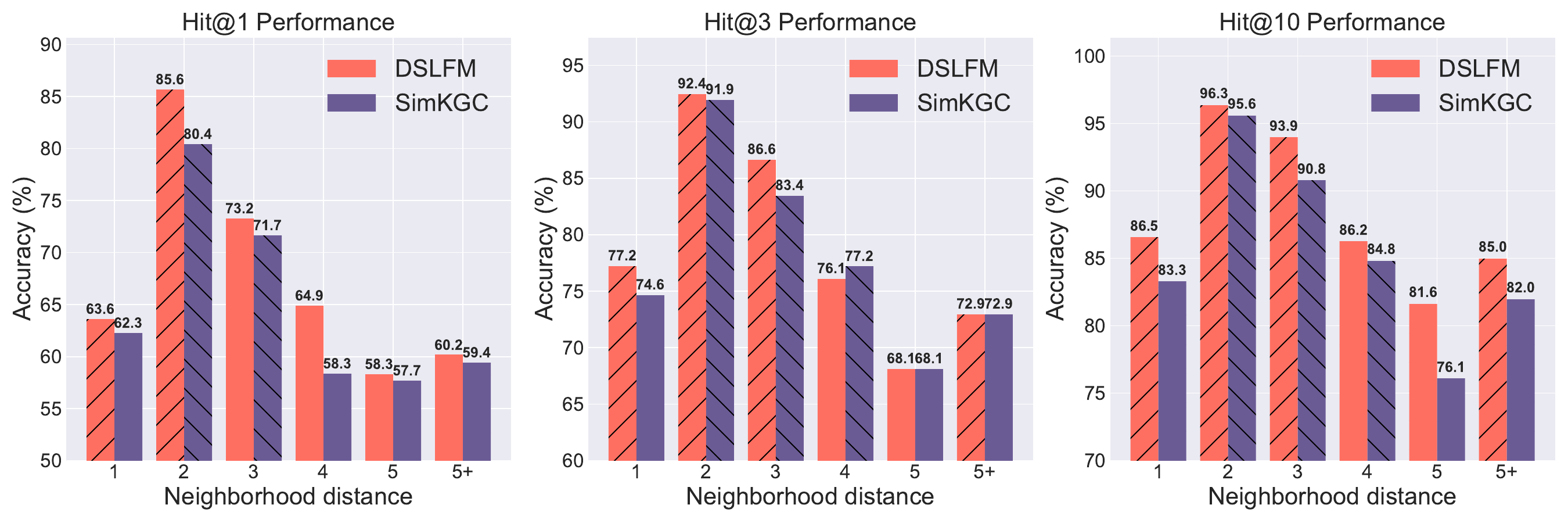}
  \vspace*{-6mm}
  \caption{Comparison of DSLFM-KGC and SimKGC performance on the WN18RR dataset \emph{w.r.t}. neighborhoods at varying geodesic distances, where ‘5+’ denotes a distance greater than 5. If no path exists between two entities, the distance is considered infinite.}
  \label{fig:analyasis_neighborhood_wn}
  \vspace*{-8mm}
\end{figure}

\textbf{Does modeling sparse community structure really benefit KGC?}. To evaluate the impact of the sparse latent structure, we replace only our posterior encoder with one that generates an approximate standard Gaussian distribution, as employed in the vanilla VAE \citep{kingma2013auto}, while keeping the BERT and MLP components unchanged. Additionally, we evaluate a pure autoencoder (AE), which does not assume a probabilistic distribution for the latent variables. The testing performance and the training convergence epochs (based on the best validation metric) for the WN18RR, FB15k-237 and UMLS datasets are presented in Table \ref{tab:ablation_latent_structure}.

Table \ref{tab:ablation_latent_structure} demonstrates that incorporating latent structure significantly improves KGC performance on the WN18RR and UMLS datasets. Furthermore, the significant performance difference observed between our model and its variants, coupled with the close results obtained by the VAE, AE, and SimKGC models, indicates that the \textbf{enhancement in KGC performance arises from the sparse community modeling itself, rather than solely from the increased model complexity}. However, on the FB15k-237 dataset, improvements are more modest, suggesting challenges in accurately capturing its underlying latent community structure. Moreover, the increased complexity introduced by the latent structure model negatively impacted convergence speed, as indicated by the longer training epochs. Future research should focus on improving KGC accuracy on densely connected KGs while simultaneously enhancing training efficiency.

\vspace*{-4.5mm}
\section{Analysis}
\vspace*{-2.5mm}
\subsection{How does clustering enhance KGC performance?}
\vspace*{-1.5mm}
To evaluate how clustering enhances KGC performance, we categorize testing triples into disjoint groups based on the geodesic distance (shortest path distance) between the head and tail entities. We then compared the Hit@k metrics for each group against those of SimKGC. The triple distributions for WN18RR and FB15k-237 are depicted in Figure \ref{fig:analyasis_neighborhood_distribution}. It is important to note that, in the Wikidata5M and UMLS test sets, no head entity within a triple is connected to any entity other than its corresponding tail entity. As demonstrated in Figure \ref{fig:analyasis_neighborhood_wn}, DSLFM-KGC exhibits superior predictive accuracy on the WN18RR dataset, especially for long-range connections. For instance, DSLFM-KGC achieves a higher Hit@1 at a distance of 4 (64.9 compared to 58.3) and an improved Hit@10 at a distance of 5 (81.6 compared to 76.1). \textbf{This observation suggests that the method effectively captures global information through community clustering, thereby enhancing KGC performance.} Results for the FB15k-237 dataset are presented in Appendix \ref{app:analysis}, where the improvement in long-range accuracy is less pronounced.

\vspace*{-5mm}
\subsection{Qualitative analysis on Interpretability}
\vspace*{-2mm}
To showcase the interpretability of our model, derived from SBM, we visualize the learned latent structure for the WN18RR and FB15k-237 datasets in Figure \ref{fig:analysis_Fmatrix}. Figure \ref{fig:analysis_Fmatrix_wn} reveals a more distinct clustering pattern in the WN18RR latent structure, characterized by larger absolute values in the left and right columns and more moderate values in the central columns. In contrast, the FB15k-237 matrix exhibits a more evenly distributed pattern of values across its columns.

Additionally, integrating a text encoder facilitates a more nuanced understanding of the latent structure learned from a KG. To illustrate this, we select several communities and their most significant entities from the FB15k-237 dataset for enhanced visualization, with their textual descriptions provided in Table \ref{tab:textual_communities_wn}. \textbf{This integration of text features enables more intuitive and concrete interpretations of the uncovered communities, thereby validating both the effectiveness and interpretability of our approach.}

\subsection{Discussion}
\vspace*{-2mm}
This section provides a comparative analysis of our method and LLM-based KGC approaches, focusing on both performance enhancement and interpretability. LLM-based methods such as CP-KGC \cite{zhang2024making} and KICGPT \cite{wei2024kicgpt} typically require extensive example demonstrations to guide LLMs toward effective KGC and mitigate hallucinations, as evidenced by the results in Table \ref{tab:kgc_results_wn_fb}. This reliance on densely connected graph contexts also extends to their interpretability. Furthermore, LLM-based methods generally struggle to scale to large KGs comprising millions of entities due to their prohibitive computational costs. However, contemporary real-world KGs are often vast and sparsely connected (or even rarely connected) \cite{wang2021kepler}, a domain in which our model demonstrates superior performance.
\vspace*{-4.5mm}
\section{Related work}
\vspace*{-3mm}

\textbf{Knowledge Graph Completion}. To address the task of KGC, initial research has concentrated on developing effective scoring mechanisms to evaluate the plausibility of triples embedded in low-dimensional spaces. A pioneering approach in this area is knowledge graph embedding (KGE) \cite{bordes2013translating, yang2014embedding, schlichtkrull2018modeling, sun2018rotate, balavzevic2019tucker}, also known as embedding-based methods. Notably, TransE \cite{bordes2013translating} is a representative model that interprets a relationship $r$ as a translation from the head entity $h$ to the tail entity $t$. Recently, text-based KGC methods have incorporated textual descriptions of entities and relations, thus encoding them into a more expressive semantic space. Specifically, NTN \cite{socher2013reasoning} simplifies entity representation by averaging its word embeddings. SimKGC \cite{wang2022simkgc} integrates a contrastive learning framework with three negative sampling strategies, significantly improving KGC performance. However, these prevalent KGC methods assume that the existence of a triple in a KG solely depends on the entities and relation involved, often overlooking the intricate interconnections among communities.

\textbf{KGC methods that leverage neighborhood information}. Graph Neural Networks (GNNs), especially Message Passing Neural Networks (MPNNs), have become essential tools for node representation learning in graphs, where they assume that similar neighborhood structures yield closer node representations. Notable MPNN-based KGC methods like RGCN \citep{schlichtkrull2018modeling}, CompGCN \citep{vashishth2019composition}, and KBGAT \citep{nathani2019learning} have demonstrated strong KGC performance but have since been found to inadequately leverage neighborhood information \citep{zhang2022rethinking, li2023message}. Furthermore, GNN-based approaches generally do not incorporate community-level information for KG completion. Meanwhile, there are few KGC methods that leverage clustering features, such as CTransR \citep{lin2015learning} and EL-Trans \citep{yang2023knowledge}. However, these models often struggle with poor KGC performance and lack an end-to-end design, limiting their applicability to modern KGs.

\textbf{Stochastic Blockmodels} have demonstrated success in uncovering various latent structures, thereby enhancing link prediction. The stochastic blockmodel (SBM) \cite{holland1983stochastic} assigns each node to a specific community, with the interconnections between nodes influenced by their community memberships. The mixed membership stochastic blockmodel (MMSB) \cite{airoldi2008mixed} introduces a multinomial indicator vector for node-community assignments, allowing for mixed membership communities. However, MMSB restricts nodes to a single cluster at any given time. The overlapping stochastic blockmodel (OSBM) \cite{latouche2011overlapping} overcomes this limitation by utilizing a multi-Bernoulli distribution, enabling nodes to belong to multiple communities simultaneously. The latent feature relational model (LFRM) \cite{miller2009nonparametric} is a specific instance of OSBM that applies the Indian Buffet Process (IBP) prior to the assignment matrix. Traditional SBMs, however, are constrained in expressiveness and scalability due to their reliance on MCMC \cite{miller2009nonparametric} or variational inference \cite{zhu2016max} for learning latent variables. Recently, DGLFRM \cite{mehta2019stochastic} employs a sparse variational autoencoder (VAE) framework for inference in SBMs, thereby extending their applicability to larger graphs. Nevertheless, DGLFRM struggles to handle graphs with tens of thousands of nodes or more, a common scenario in modern KGs.

\vspace*{-5mm}
\section{Conclusion}
\vspace*{-3mm}
In this paper, we introduce {\modelname}, a novel framework designed to learn sparse latent structural features for improved knowledge graph completion (KGC). Our approach leverages a probabilistic model for KGs inspired by stochastic blockmodels (SBMs), which dynamically discovers latent communities within the graph and subsequently enhances triple completion accuracy. To facilitate scalable inference and enhance model expressiveness, we incorporate a deep sparse variational autoencoder. Through extensive experimentation on three benchmark datasets, we demonstrate the superior performance of {\modelname} in completing missing triples while maintaining interpretability. Despite these advancements, optimizing training efficiency remains a critical challenge. Future research will focus on developing methods to learn more expressive latent representations while simultaneously reducing computational overhead.

\bibliographystyle{plain}
\bibliography{ref}

\newpage
\appendix
\section{Clustering Benefits for KGC: An Example}
\begin{figure}[h]
  \centering
  \includegraphics[width=.8\textwidth]{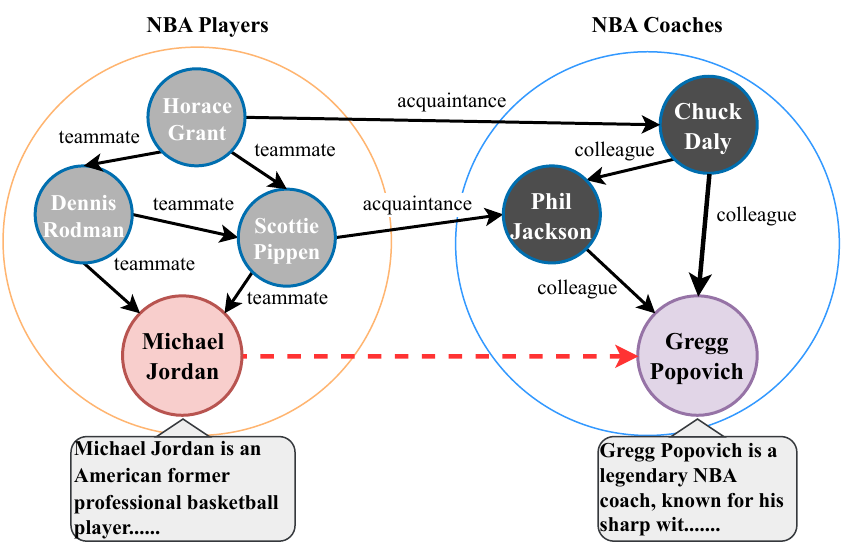}
  \caption{A simplified example of KG involving diverse communities. Solid black arrows indicate existing links, while the dashed red arrow represents a missing link for a KGC model to predict. Each community within the graph is encircled, highlighting the overlapped groups of interconnected entities.}
  \label{fig:introduction}
\end{figure}

In the KG illustrated in Figure \ref{fig:introduction}, entities are organized into overlapping communities. To address the query concerning the relationship between Michael Jordan and Gregg Popovich, "acquaintance" presents itself as a likely candidate, inferable from the observed interconnections between the "NBA Players" and "NBA Coaches" communities. This inference is not readily attainable through textual information alone.

\section{Mathematical proofs}\label{app:math}
This section provides detailed mathematical derivations of the negative ELBO as introduced in Equation \ref{eq:elbo}.

\subsection{The negative ELBO}
Let $\mathcal{H} = \{V_\text{qry}, Z_\text{qry}, W_\text{qry}, V_\text{ans}, Z_\text{ans}, W_\text{ans}\}$ denote the set of latent variables and $\mathcal{O} = \{X_\text{qry}, X_\text{ans}, A\}$ the set of observations, with $X_\text{qry}$ and $X_\text{ans}$ being the tokenized sequences of the queries and answer, respectively. The negative ELBO in our model is formulated as:
\begin{align*}
\mathcal{L}
& = -\mathbb{E}_q \left[ \log \frac{p_\theta(\mathcal{H}, \mathcal{O})}{q_\phi(\mathcal{H})} \right] \\
& = -\mathbb{E}_q \left[ \log \frac{p_\theta(\mathcal{H})}{q_\phi(\mathcal{H})} + \log p_\theta(\mathcal{O} |\mathcal{H}) \right] \\
& = \KL\left[ q_\phi(\mathcal{H}) || p_\theta(\mathcal{H}) \right] - \mathbb{E}_q \left[ \log p_\theta(\mathcal{O} | Z_\text{qry}, Z_\text{ans}, W_\text{qry}, W_\text{ans}) \right] \\
& = \KL[q_\phi(\mathcal{H}_\text{qry}) || p_\theta(\mathcal{H}_\text{qry})] + \KL[q_\phi(\mathcal{H}_\text{ans}) || p_\theta(\mathcal{H}_\text{ans})] - \mathbb{E}_q \left[ \log p_\theta(X_\text{qry}, X_\text{ans}, A | Z_\text{qry}, Z_\text{ans}, W_\text{qry}, W_\text{ans}) \right] \\
& = \KL[q_\phi(\mathcal{H}_\text{qry}) || p_\theta(\mathcal{H}_\text{qry})] + \KL[q_\phi(\mathcal{H}_\text{ans}) || p_\theta(\mathcal{H}_\text{ans})] - \mathbb{E}_q \left[ \log p_\theta(X_\text{qry} | Z_\text{qry}, W_\text{qry}) \right] \\
& ~~ - \mathbb{E}_q \left[ \log p_\theta(X_\text{ans} | Z_\text{ans}, W_\text{ans}) \right] - \mathbb{E}_q \left[ \log p_\theta(A | Z_\text{qry}, Z_\text{ans}, W_\text{qry}, W_\text{ans}) \right]
\end{align*}

This objective consists of three main parts: the first two KL divergence terms $\mathcal{L}_\text{KL}$, the next two reconstruction terms $\mathcal{L}_\text{recon}$, and the last triple completion term $\mathcal{L}_\text{comp}$.

Given a batch of triples $B \subset \mathcal{G}$, with $\mathcal{Q}_B$ and $\mathcal{E}_B$ representing the associated queries and answers, we derive a batch-optimized version of Equation \ref{eq:elbo}:
{\begin{align}\label{eq:elbo_batch}
\mathcal{L}(B)
& = \mathcal{L}_\text{KL}(B) + \mathcal{L}_\text{Recon}(B) + \mathcal{L}_\text{Comp}(B) \notag \\
& = \sum_{(h,r) \in \mathcal{Q}_B} \left\{ \KL\left[ q_\phi(\mathbf{v}_{hr}) || p_\theta(\mathbf{v}_{hr}) \right] + \KL\left[ q_\phi(\mathbf{z}_{hr}) || p_\theta(\mathbf{z}_{hr} | \mathbf{v}_{hr}) \right] + \KL\left[ q_\phi(\mathbf{w}_{hr}) || p_\theta(\mathbf{w}_{hr}) \right] \right\} \notag \\
& ~~ + \sum_{t \in \mathcal{E}_B} \left\{ \KL\left[ q_\phi(\mathbf{v}_t) || p_\theta(\mathbf{v}_t) \right] + \KL\left[ q_\phi(\mathbf{z}_t) || p_\theta(\mathbf{z}_t | \mathbf{v}_t) \right] + \KL\left[ q_\phi(\mathbf{w}_t) || p_\theta(\mathbf{w}_t) \right] \right\} \notag \\
& ~~ - \sum_{(h,r) \in \mathcal{Q}_B} \mathbb{E}_q \left[ \log p_\theta(\mathbf{x}_{hr} | \mathbf{z}_{hr}, \mathbf{w}_{hr}) \right] - \sum_{t \in \mathcal{E}_B} \mathbb{E}_q \left[ \log p_\theta(\mathbf{x}_{t} | \mathbf{z}_{t}, \mathbf{w}_{t}) \right] \notag \\
& ~~ - \sum_{(h,r,t) \in B} \mathbb{E}_q \left[ \log p_\theta(A_{hr,t} | \mathbf{z}_{hr}, \mathbf{z}_{t}, \mathbf{w}_{hr}, \mathbf{w}_{t}) \right]
\end{align}}

To compute the reconstruction terms, we use the cosine similarity between the embedding $\mathbf{e}_{hr}$ (Equation \ref{eq:bert_encoding}) and the decoded representation $\mathbf{g}_{hr}$ (Equation \ref{eq:mlp_decoding}).

Note that, $\mathbf{v}_{hr}, \mathbf{v}_{t}, \mathbf{z}_{hr}, \mathbf{z}_{t}, \mathbf{w}_{hr}$ and are $K$-dimensional vectors, while $\mathbf{g}_{hr}, \mathbf{g}_{t} \in \mathbb{R}^{D}$. The time required to compute 
$\mathcal{L}_\text{KL}(B)$, $\mathcal{L}_\text{Recon}(B)$ and $\mathcal{L}_\text{Comp}(B)$ in Equation $\ref{eq:elbo_batch}$ is $\mathcal{O}(|B| \cdot C_\text{KL})$, $\mathcal{O}(|B| \cdot C_\text{Recon})$ and $\mathcal{O}(|B| \cdot C_\text{Comp})$, with space complexities $\mathcal{O}(|B| \cdot K)$, $\mathcal{O}(|B| \cdot D)$ and $\mathcal{O}(|B| \cdot D)$, respectively. Here, $C_\text{KL}$, $C_\text{Recon}$ and $C_\text{Comp}$ denote the complexity for evaluating the KL divergence, reconstruction and triple completion terms for a single triple. Thus, the total time and space complexity for computing \ref{eq:elbo_batch} are $\mathcal{O}(|B| \cdot (C_\text{KL} + C_\text{Recon} + C_\text{Comp}))$ and $\mathcal{O}(|B| \cdot D + |B| \cdot K)$.

In practice, we apply two different weighting coefficients to the KL and reconstruction losses to balance the learning objectives and reduce the risk of posterior collapse \citep{higgins2017beta}:
\begin{equation}
\mathcal{L}(B) = \beta\mathcal{L}_\text{KL}(B) + \eta\mathcal{L}_\text{Recon}(B) + \mathcal{L}_\text{Comp}(B)
\end{equation}

Regarding the KL terms, we adhere to the method described by \cite{kingma2013auto} for computing the KL divergences for two normal variables: $\KL\left[ q_{\phi}(\mathbf{w}_{hr}) || p_\theta(\mathbf{w}_{hr}) \right]$ and $\KL\left[ q_{\phi}(\mathbf{w}_{t}) || p_\theta(\mathbf{w}_{t}) \right]$. In the following sections, we derive the computation of the KL divergence for two Beta distributions, \emph{i.e}., $\KL\left[ q_{\phi}(\mathbf{v}_{hr}) || p_\theta(\mathbf{v}_{hr}) \right]$ and $\KL\left[ q_{\phi}(\mathbf{v}_{t}) || p_\theta(\mathbf{v}_{t}) \right]$, as well as for two Bernoulli distributions, \emph{i.e}., $\KL\left[ q_{\phi}(\mathbf{z}_{hr}) || p_\theta(\mathbf{z}_{hr} | \mathbf{v}_{hr}) \right]$ and $\KL\left[ q_{\phi}(\mathbf{z}_{t}) || p_\theta(\mathbf{z}_{t} | \mathbf{v}_{t}) \right]$.

\subsection{The KL divergence of Beta distributions}
The KL divergence of two Beta distributions has a closed-form solution. The PDF of a Beta distribution $\text{Beta}(a, b)$ with concentration parameters $a, b$ is given by:
\begin{equation}
f(x | a, b) = \frac{1}{\text{B}(a, b)} x^{a-1} (1-x)^{b-1}, \quad 0 \leq x \leq 1
\end{equation}
where $\text{B}(a, b)$ is the Beta function, defined as
\begin{equation}
\text{B}(a, b) = \int_0^1 u^{a-1} (1-u)^{b-1} du = \frac{\Gamma(a)\Gamma(b)}{\Gamma(a+b)}
\end{equation}
with $\Gamma(\cdot)$ representing the Gamma function.

Let the distributions $p(x)$ and $q(x)$ be $\text{Beta}(a_p, b_p)$ and $\text{Beta}(a_q, b_q)$ respectively, the KL divergence for $q(x)$ and $p(x)$ is computed as:
\begin{align*}
KL\left[q(x) || p(x)\right]
& = \mathbb{E}_q \left[ \log\frac{q(x)}{p(x)} \right] \\
& = \mathbb{E}_q \left[ \log\frac{\frac{1}{\text{B}(a_q, b_q)} x^{a_q-1}(1-x)^{b_q-1}}{\frac{1}{\text{B}(a_p, b_p)} x^{a_p-1}(1-x)^{b_p-1})} \right] \\
& = \mathbb{E}_q \left[\log\frac{\text{B}(a_p, b_p)}{\text{B}(a_q, b_q)} \right] + (a_q-a_p)\mathbb{E}_q\left[ \log x \right] + (b_q-b_p)\mathbb{E}_q\left[ \log(1-x) \right] \\
& = \log\text{B}(a_p, b_p) - \log\text{B}(a_q, b_q) + (a_q-a_p)\mathbb{E}_q\left[ \log x \right] ~~ + (b_q-b_p)\mathbb{E}_q\left[ \log(1-x) \right]
\end{align*}

where $\mathbb{E}_q[\log x]$ and $\mathbb{E}_q[\log (1-x)]$ are the expected sufficient statistics under distribution $q$, which can be computed using the properties of the exponential family distributions:
\begin{align}
\mathbb{E}_q[\log x] & = \psi(a_q) - \psi(a_q+b_q) \\
\mathbb{E}_q[\log (1-x)] & = \psi(b_q) - \psi(a_q+b_q)
\end{align}
where $\psi(\cdot)$ denotes the di-gamma function.

Thus, the complete expression of the KL divergence becomes:
\begin{align}
KL\left[q(x) || p(x)\right]
& = \log\text{B}(a_p, b_p) - \log\text{B}(a_q, b_q) + (a_q-a_p)(\psi(a_q) - \psi(a_q+b_q)) \notag \\
& \quad + (b_q-b_p)(\psi(b_q) - \psi(a_q+b_q))
\end{align}

\subsection{The KL divergence of Concrete distributions}
To enable differentiable optimization, we utilize the binary Concrete distribution \citep{maddison2016concrete} to obtain a continuous relaxation of the Bernoulli distribution (Equation \ref{eq:q_zk}). However, the KL divergence of two Concrete distributions, $q(y)$ and $p(y)$, is intractable. We resort to approximation using the Monte Carlo (MC) expectations:
\begin{align}\label{eq:kl_concrete}
KL\left[q(y) || p(y)\right]
& = \mathbb{E}_q \left[ \log q(y) - \log p(y) \right] \notag \\
& \simeq \frac{1}{N} \sum_{i=1}^N (\log q(y_i) - \log p(y_i)), \quad y_i \sim q(y), \ i=1,\ldots,N
\end{align}

According to \cite{maddison2016concrete}, the logarithm of the probability density function for the Concrete distribution is given by:
\begin{equation}
\log p(y|\pi,\lambda) = \log\lambda - \lambda y + \log\pi - 2\log(1+\exp(-\lambda y + \log\pi))
\end{equation}
where $p(y|\pi,\lambda) \triangleq \text{Concrete}(y|\pi,\lambda)$ denotes the Concrete distribution, $\lambda \in (0,\infty)$ the relaxation temperature, and $\pi$, the probability ratio.

Specifically, the KL divergence term $\KL\left[ q_{\phi}(\mathbf{z}_{hr}) || p_\theta(\mathbf{z}_{hr} | \mathbf{v}_{hr}) \right]$ in the negative ELBO (Equation \ref{eq:elbo}) is computed as:
\begin{align}
\KL\left[ q_{\phi}(\mathbf{z}_{hr}) || p_\theta(\mathbf{z}_{hr} | \mathbf{v}_{hr}) \right]
& = \mathbb{E}_q\left[ \log q_{\phi}(\mathbf{z}_{hr}) - \log p_\theta(\mathbf{z}_{hr} | \mathbf{v}_{hr})  \right] \notag \\
& = \sum_{k=1}^K \mathbb{E}_q [\log q_{\phi}({z}_{hr,k}) - \log p_\theta({z}_{hr,k} | \mathbf{v}_{hr})]
\end{align}
where we apply the Concrete relaxation to the variational posterior (Equation \ref{eq:q_zk}) and the prior (Equation \ref{eq:pi_z}):
\begin{align}
q_{\phi}({z}_{hr,k}) & \triangleq \text{Concrete}({z}_{hr,k} | \pi_{hr,k}(\mathcal{G}), \lambda_\text{post}) \\
p_{\theta}({z}_{hr,k} | \mathbf{v}_{hr}) & \triangleq \text{Concrete}({z}_{hr,k} | \pi_{hr,k}(\mathbf{v}_{hr}), \lambda_\text{prior})
\end{align}

In this case, $\lambda_\text{post}$ and $\lambda_\text{prior}$ are hyperparameters and we have
\begin{equation}
\pi_{hr,k}(\mathbf{v}_{hr}) = \prod_{j=1}^k v_{hr,j}, ~v_{hr,j} \sim q_{\phi}(v_{hr,j})
\end{equation}
where $q_{\phi}(v_{hr,j})$ is defined in Equation \ref{eq:q_vk}. Then $\KL\left[ q_{\phi}(\mathbf{z}_{hr}) || p_\theta(\mathbf{z}_{hr} | \mathbf{v}_{hr}) \right]$ is estimated using Equation \ref{eq:kl_concrete}. The computation of $\KL\left[ q_{\phi}(\mathbf{z}_{hr}) || p_\theta(\mathbf{z}_{hr} | \mathbf{v}_{hr}) \right]$ is implemented similarly.

\subsection{Reparameterization}\label{app:reparam}
In our model, the expectations over Beta, Bernoulli and Normal distributions is approximated using differentiable Monte Carlo (MC) estimate, as required by SGVB \citep{kingma2013auto}. Furthermore, to draw samples from these distributions, a reparameterization trick is needed to ensure effective differentiation. To sample Normal variables $\mathbf{w}_{hr}$ and $\mathbf{w}_t$, we follow the standard approach used in vanilla VAE \citep{kingma2013auto}. For reparameterization of Beta variables $\mathbf{v}_{hr}$ and $\mathbf{v}_t$, we adopt the implicit differentiation method \citep{figurnov2018implicit}.

To draw discrete Bernoulli variables during training, we utilize the Gumbel-max relaxation \citep{maddison2014sampling, jang2016categorical} to achieve a continuous approximation. Specifically, the distribution used to reparameterize $z_{hr,k}$ aligns with a binary special case of the Concrete distribution \citep{maddison2016concrete}:
\begin{gather}
u \sim \text{Uniform}(0,1) \quad L = \log(u) - \log(1-u) \notag \\
y_{hr,k} \mathop{=}\limits^{d} (\text{logit}(\pi_{hr,k}) + L)/ \lambda \notag \\
z_{hr,k} = \sigma\left( y_{hr,k} \right)
\end{gather}
where $z_{hr,k}$ and $\pi_{hr,k}$ are defined in Equation \ref{eq:q_zk} and \ref{eq:mlp}, $\text{logit}(\cdot)$ is the inverse-sigmoid function and $\lambda$ is the relaxation temperature. The reparameterization of $z_{t,k}$ is achieved similarly.

\section{Experimental Details}\label{app:experiment_settings}
\vspace*{-3.5mm}
\begin{table*}[t]
  \caption{Statistics of datasets.}
  \centering
  \begin{tabular}{ccccccc}
    \toprule[1pt]
    \midrule
    Dataset & \# Relation & \# Entity & \# Triple & \# Train & \# Validation & \# Test \\
    \midrule
    UMLS & 46 & 135 & 6,556 & 5,327 & 596 & 633 \\
    WN18RR & 11 & 40,943 & 93,003 & 86,835 & 3,034 & 3,134 \\
    FB15k-237 & 237 & 14,541 & 310,116 & 272,115 & 17,535 & 20,466 \\
    Wikidata5M & 822 & 4,594,485 & 20,624,605 & 20,614,279 & 5,163 & 5,163 \\
    \midrule
    \bottomrule[1pt]
  \end{tabular}
  \label{tab:data-statistics}
\end{table*}

\begin{table*}[t]
  \caption{Hyperparameters of our \modelname ~for each dataset during training.}
  \centering
  \begin{tabular}{c|ccc}
    \toprule[1pt]
    \midrule
    Hyperparameter & WN18RR & FB15k-237 & Wikidata5M \\
    \midrule
    initial learning rate & $8 \times 10^{-5}$ &$2 \times 10^{-5}$ & $5 \times 10^{-5}$ \\
    epochs & 65 & 15 & 1 \\
    contrastive temperature $\tau$ & 0.02 & 0.08 & 0.03 \\
    dropout & 0 & 0.1 & 0 \\
    stick-breaking prior $\alpha_\text{qry}$ & 100 & 100 & 100 \\
    stick-breaking prior $\alpha_\text{ans}$ & 20 & 20 & 100 \\
    truncation level $K$ & 128 & 128 & 128 \\
    \midrule
    \bottomrule[1pt]
  \end{tabular}
  \label{tab:hyperparameters}
  \vspace*{-3.5mm}
\end{table*}

\begin{figure}[h]
  \centering
    \subfloat[WN18RR]{
    \includegraphics[width=0.4\linewidth]{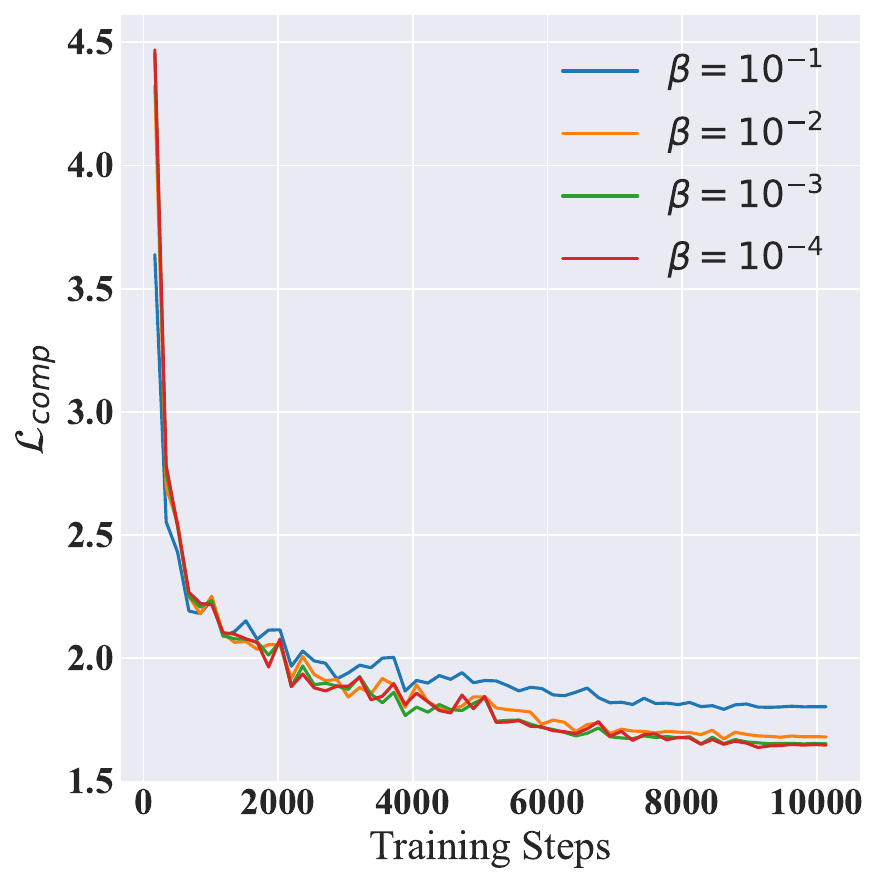}
    \label{fig:ablation_klweight_wn}
    }
    % \hfill
    \subfloat[FB15k-237]{
        \includegraphics[width=0.4\linewidth]{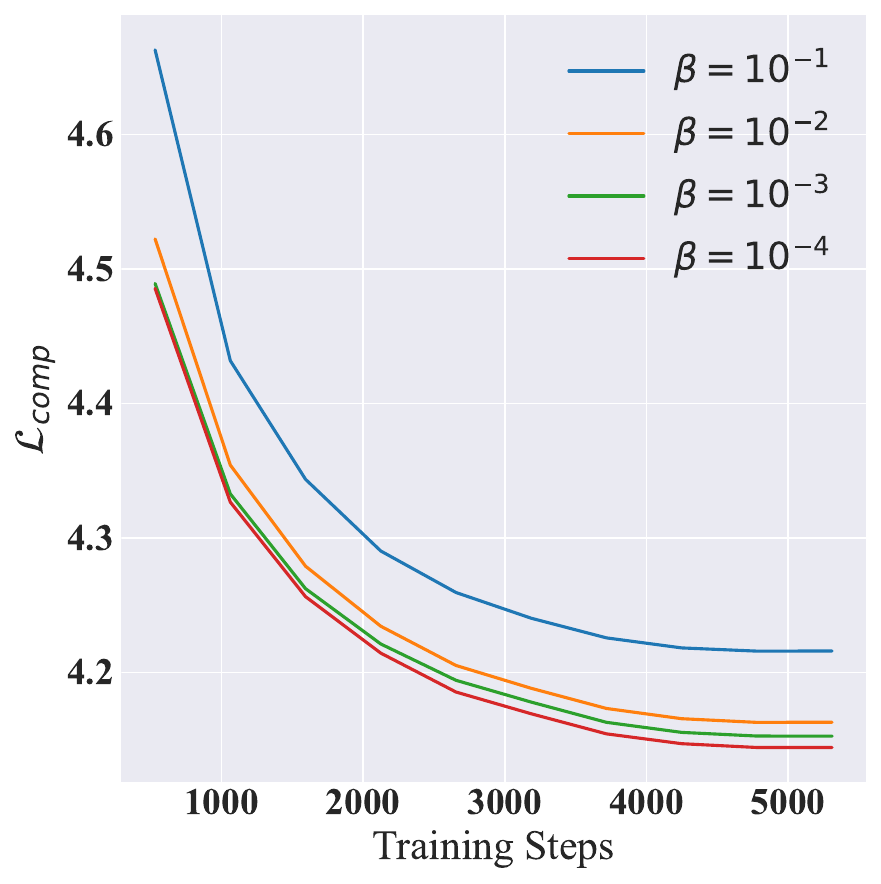}
        \label{fig:ablation_klweight_fb}
    }
  \caption{Validation triple completion loss $\mathcal{L}_\text{comp}$ for {\modelname} during training with different $\beta$ values on the WN18RR and FB15k-237 datasets.}
  \label{fig:ablation_klweight}
\end{figure}

\textbf{Datasets}. The statistics of each dataset are shown in Table \ref{tab:data-statistics}. The UMLS encompasses a wide array of biomedical concepts, including names for drugs and diseases, alongside the relationships that exist between them, such as diagnosis and treatment. Originally introduced by \citep{bordes2013translating}, the WN18 and FB15k datasets were later refined to WN18RR and FB15k-237 following studies \citep{toutanova2015representing, dettmers2018convolutional} that revealed test leakage issues. Textual data comes from KG-BERT \citep{yao2019kg}. Wikidata5M-Ind \citep{wang2021kepler} is a large-scale KG merging Wikidata and Wikipedia, with textual descriptions for each entity.

\begin{table}[t]
    \centering
    \caption{The parameter count, training epochs, and GPU hours required by SimKGC \citep{wang2022simkgc} and DSLFM-KGC.}
    \label{tab:compute_overhead}
    \resizebox{\textwidth}{!}{\begin{tabular}{c|c|cc|cc|cc}
        \toprule[1pt]
        \midrule
        \multirow{2}{*}{Model} & \multirow{2}{*}{\# Params} & \multicolumn{2}{c|}{WN18RR} & \multicolumn{2}{c|}{FB15k-237} & \multicolumn{2}{c}{Wikidata5M} \\
        & & Epochs & GPU hours & Epochs & GPU hours & Epochs & GPU hours \\
        \midrule
        SimKGC & 218.0M & 50 & 3 & 10 & 2 & 1 & 12 \\
        DSLFM-KGC (ours) & 219.8M & 65 & 3.5 & 15 & 3 & 1 & 13 \\
        \midrule
        \bottomrule[1pt]
  \end{tabular}}
\end{table}

\begin{table}[t]
    \caption{\footnotesize{Performance of {\modelname} on the WN18RR, FB15k-237 and Wikidata5M datasets w/ different stick-breaking priors.} }
    \centering
    \label{tab:ablation_alpha}
    \resizebox{.8\linewidth}{!}{
      \begin{tabular}{c|cc|cc|cc}
        \toprule[1pt]
        \midrule
         \multirow{2}{*}{Dataset}& \multicolumn{2}{c|}{WN18RR} & \multicolumn{2}{c|}{FB15k-237} & \multicolumn{2}{c}{Wikidata5M} \\
        & Hit@1 & Hit@10 & Hit@1 & Hit@10 & Hit@1 & Hit@10 \\
        \midrule
        Mean & 62.7 & 84.1 & 26.2 & 53.8 & 66.9 & 94.0 \\
        \midrule
        Std & 0.3 & 0.1 & 0.2 & 0.1 & 0.4 & 0.2 \\
        \midrule
        \bottomrule[1pt]
      \end{tabular}
    }
    \vspace*{-3.5mm}
\end{table}

\textbf{Baselines}. The baseline methods we choose can be categorized into four classes:
\begin{itemize}
  \item In the category of embedding-based methods, we choose TransE \citep{bordes2013translating}, DistMult \citep{yang2014embedding}, ConvE \citep{dettmers2018convolutional}, RotatE \citep{sun2018rotate}, TuckER \citep{balavzevic2019tucker}, HittER \citep{chen-etal-2021-hitter} and KRACL \cite{tan2023kracl}.
  \item For GNN-based methods, we consider R-GCN \citep{schlichtkrull2018modeling}, CompGCN \cite{vashishth2019composition} and KGGAT \cite{nathani2019learning}.
  \item Text-based methods considered include KG-BERT \citep{yao2019kg}, MTL-KGC \citep{kim-etal-2020-multi}, StAR \citep{wang2021structure}, KG-S2S \citep{chen2022knowledge}, DKPL \citep{xie2016representation}, KEPLER \citep{wang2021kepler}, BLP \cite{daza2021inductive}, SimKGC \citep{wang2022simkgc} and GHN \citep{qiao-etal-2023-improving}.
  \item For the LLM-based methods, we include CP-KGC \cite{yang2024enhancing} and KICGPT \cite{wei2024kicgpt}.
\end{itemize}

\textbf{Implementation details}. We utilize two separate BERT encoders to process the textual descriptions of the queries and answers. For a specific query $(h,r)$ and entity $t$, the token sequences, \emph{i.e.}, $\mathbf{x}_{hr}$ and $\mathbf{x}_{t}$, are defined as follows:

\vspace*{-4mm}
\begin{align}
x_{hr} & = \left[ \text{CLS}, \mathcal{M}(h), \text{SEP}, \mathcal{M}(r), \text{SEP} \right] \\
x_{t} & = \left[ \text{CLS}, \mathcal{M}(t), \text{SEP} \right] 
\end{align}
where $\text{CLS}$ and $\text{SEP}$ are special tokens introduced by \cite{devlin-etal-2019-bert}, and $\mathcal{M}(h), \mathcal{M}(r)$, and $\mathcal{M}(t)$ represent the tokenized textual descriptions of the head, relation and tail, respectively. Following tokenization, $\mathbf{x}_{hr}$ and $\mathbf{x}_{t}$ are processed through BERT encoders, as specified in Equation \ref{eq:bert_encoding}.

\textbf{Hyerparameter}. Table \ref{tab:hyperparameters} lists the consistent hyperparameters used for each dataset.

\begin{figure}[t]
  \centering
  \includegraphics[width=.7\linewidth]{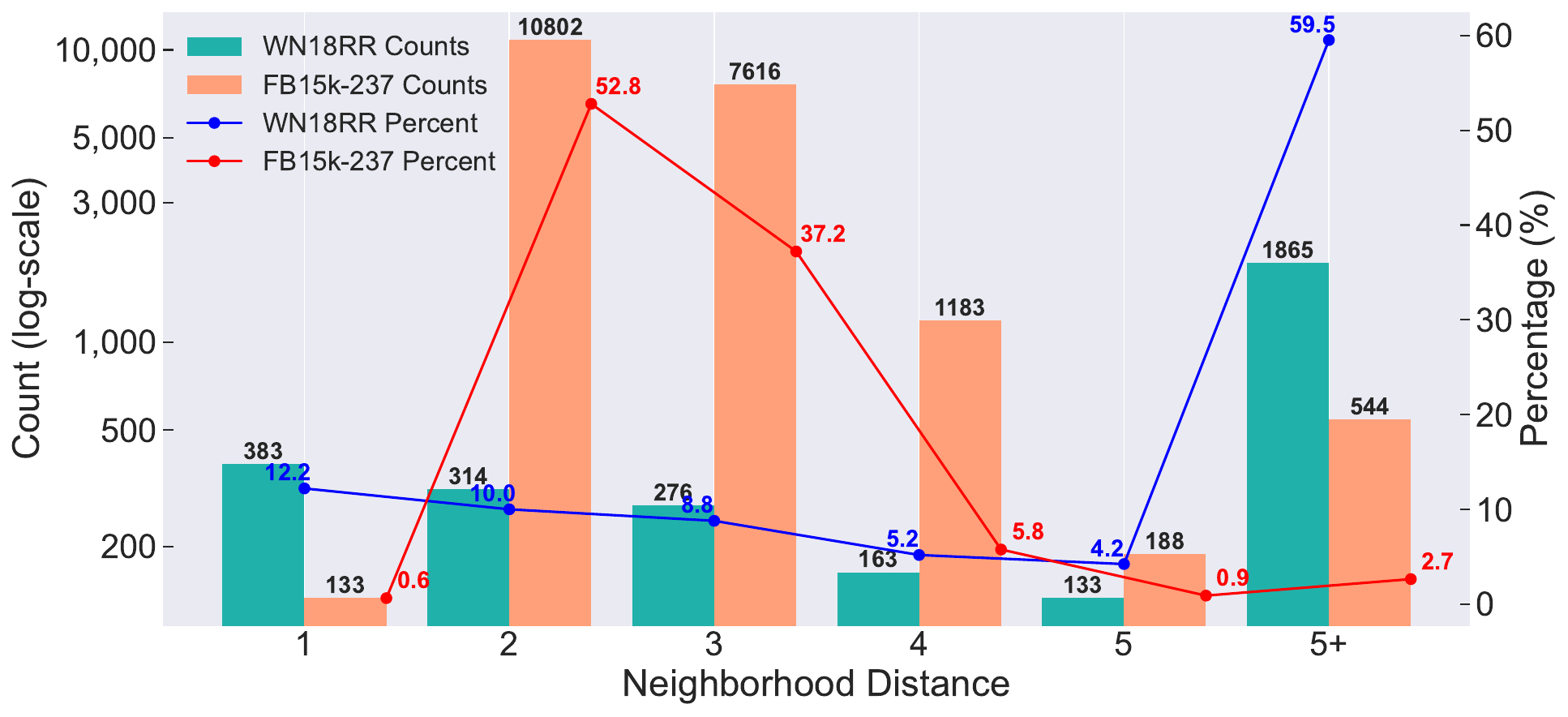}
  \vspace*{-2mm}
  \caption{Testing data distribution of the WN18RR and FB15k-237 dataset \emph{w.r.t.} neighborhoods at varying geodesic distances, where ‘5+’ indicates a distance greater than 5.}
  \label{fig:analyasis_neighborhood_distribution}
  \vspace*{-3mm}
  \end{figure}

\begin{figure}[t]
  \centering
  \includegraphics[width=1\linewidth]{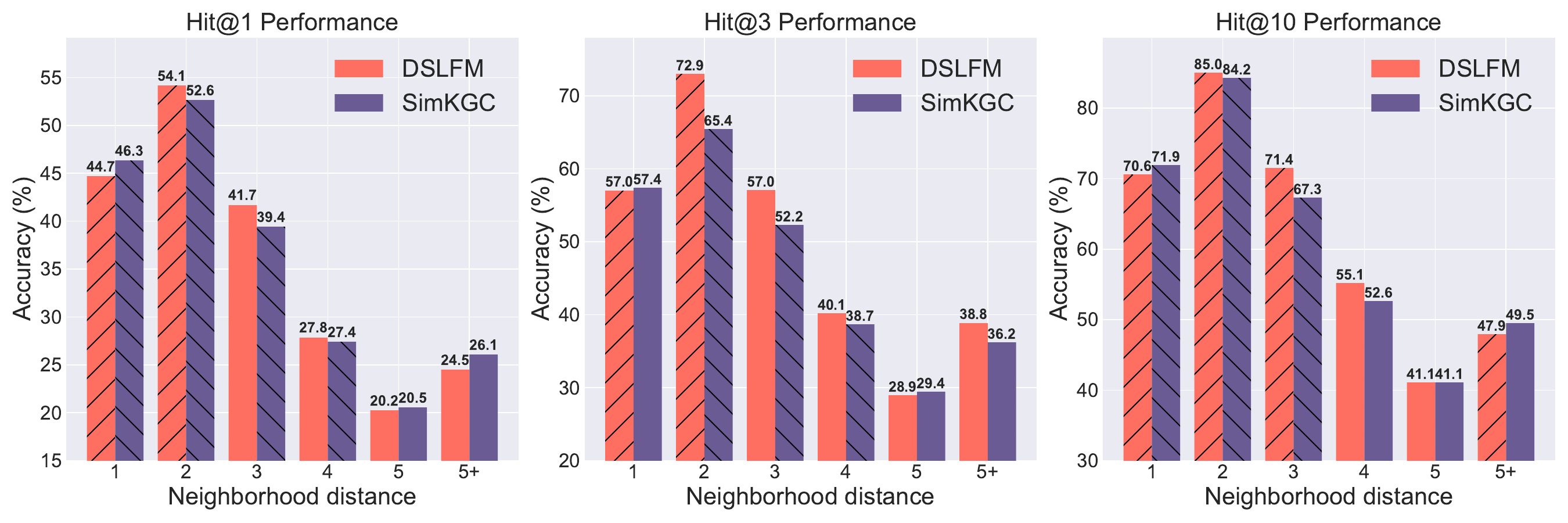}
  \caption{Comparison of DSLFM-KGC and SimKGC performance on the FB15k-237 dataset \emph{w.r.t}. neighborhoods at varying geodesic distances, where ‘5+’ denotes a distance greater than 5. If no path exists between two entities, the distance is considered infinite.}
  \label{fig:analyasis_neighborhood_fb}
\end{figure}

\section{Additional results}
\begin{wrapfigure}{r}{0.6\textwidth}
    \vspace*{-4.5mm}
    \centering
    \caption{Performance of {\modelname} on the WN18RR and FB15k-237 datasets w/ different $\beta$ values.}
    \label{tab:ablation_klweight}
    \resizebox{\linewidth}{!}{
      \begin{tabular}{c|ccc|ccc}
        \toprule[1pt]
        \midrule
        \multirow{2}{*}{$\beta$} & \multicolumn{3}{c|}{WN18RR} & \multicolumn{3}{c}{FB15k-237} \\
        & MRR & Hit@1 & Hit@10 & MRR & Hit@1 & Hit@10 \\
        \midrule
        $10^{-1}$ & 69.2 & 61.6 & 83.3 & 33.7 & 24.5 & 52.2 \\
        $10^{-2}$ & 70.2 & 62.8 & 83.9 & 35.1 & 26.0 & 53.3 \\
        $10^{-3}$ & 70.2 & 62.6 & 84.3 & 35.4 & 26.2 & 53.6 \\
        $10^{-4}$ & 70.4 & 62.5 & 84.0 & 35.4 & 26.2 & 53.7 \\
        \midrule
        \bottomrule[1pt]
      \end{tabular}
      }
      \vspace*{-3.5mm}
\end{wrapfigure}

\subsection{Additional ablation results}\label{app:ablation}
Figure \ref{fig:ablation_klweight} and Table \ref{tab:ablation_klweight} depict the training behavior and testing performance of {\modelname} across various KL weight $\beta$ settings. For both the WN18RR and FB15k-237 datasets, setting $\beta=10^{-1}$ leads to a learning imbalance between the KL and triple completion losses, which negatively impacts the validation loss. In contrast, the validation loss ($\mathcal{L}_\text{comp}$) curves for {\modelname} with $\beta=10^{-2}, 10^{-3},$ and $10^{-4}$ show minimal variation. This observation is mirrored in the testing results shown in Table \ref{tab:ablation_klweight}.

\subsection{Additional analysis results}\label{app:analysis}
\begin{wrapfigure}{r}{0.3\textwidth}
  \vspace*{-4.5mm}
  \caption{Modularity score of the WN18RR and FB15k-237 datasets.}
  \centering
  \resizebox{\linewidth}{!}{
    \begin{tabular}{cc}
      \toprule[1pt]
      \midrule
      Dataset & Modularity score \\
      \midrule
      WN18RR & 0.574 \\
      FB15k-237 & 0.074 \\
      \midrule
      \bottomrule[1pt]
    \end{tabular}
  }
  \label{tab:modularity}
  \vspace*{-3mm}
\end{wrapfigure}

\textbf{Modularity of the KGs}. To assess the significance of clustering within the WN18RR and FB15k-237 datasets, we quantified the modularity of each. Specifically, we applied the label propagation algorithm for community detection and subsequently calculated the modularity score of the resulting partitions using the following equation:
\begin{equation}
Q = \frac{1}{2m} \sum_{i,j} \left( A_{ij} - \gamma\frac{k_i k_j}{2m} \right) \delta(c_i, c_j)
\end{equation}
where $m$ is the number of edges, $A$ is the adjacency matrix, $\gamma$ is the resolution parameter, $k_i$ and $k_j$ are the degrees of nodes $i$ and $j$, and $\delta(c_i, c_j)$ is 1 if nodes $i$ and $j$ belong to the same community, and 0 otherwise. A modularity score closer to 1 indicates a more distinct clustering structure.

\textbf{Geodesic analysis on the FB15k-237 dataset}. Figure \ref{fig:analyasis_neighborhood_fb} presents the performance comparison between DSLFM-KGC and SimKGC \emph{w.r.t}. geodesic distance on the FB15k-237 dataset. To determine the triple distribution, we first construct the graph using training data and subsequently compute the shortest path distance between the head and tail entity for each triple.

\begin{figure}[h!]
  \centering
    \subfloat[WN18RR]{
    \includegraphics[width=0.4\linewidth]{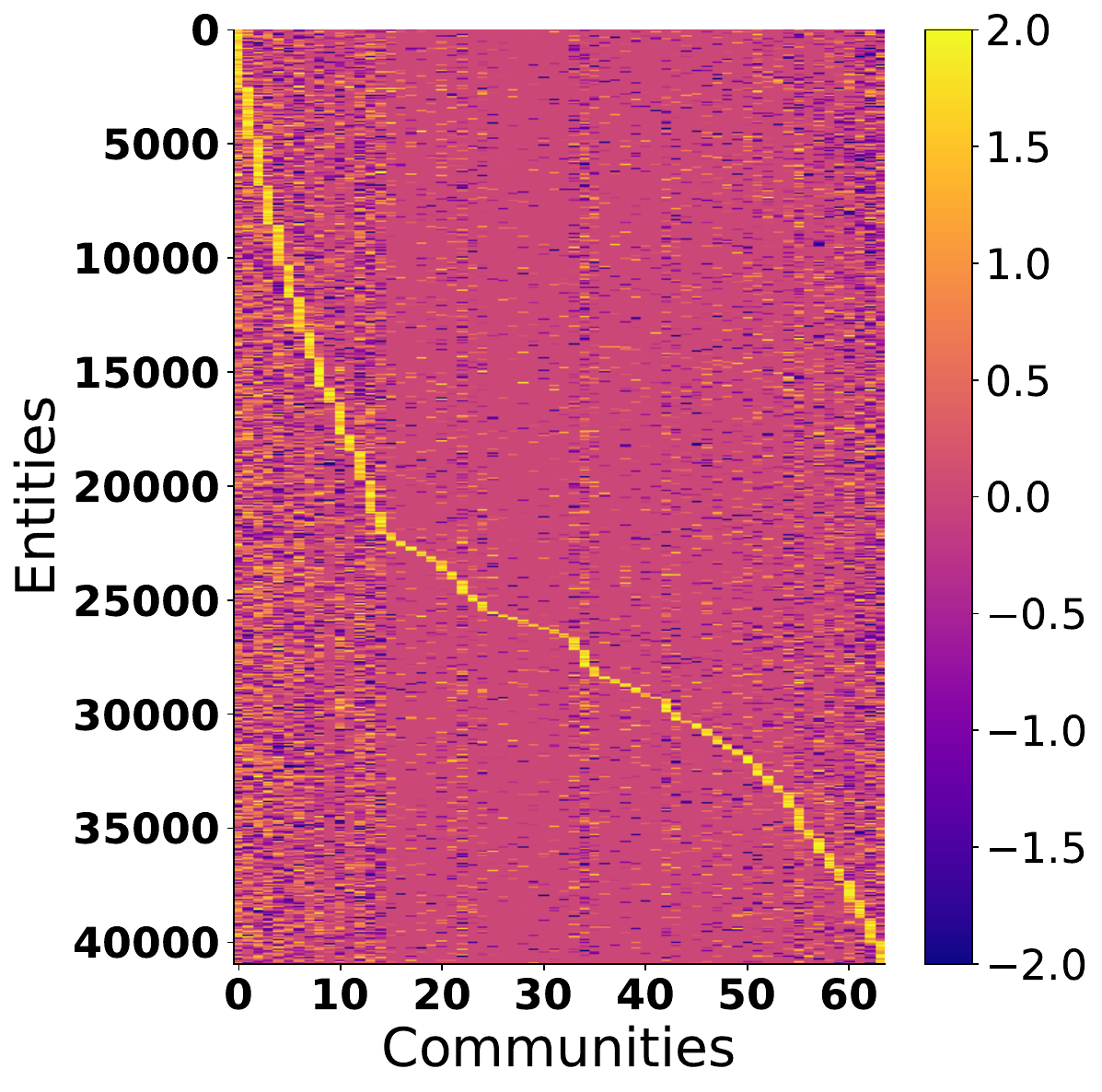}
    \label{fig:analysis_Fmatrix_wn}
    }
    % \hfill
    \subfloat[FB15k-237]{
        \includegraphics[width=0.4\linewidth]{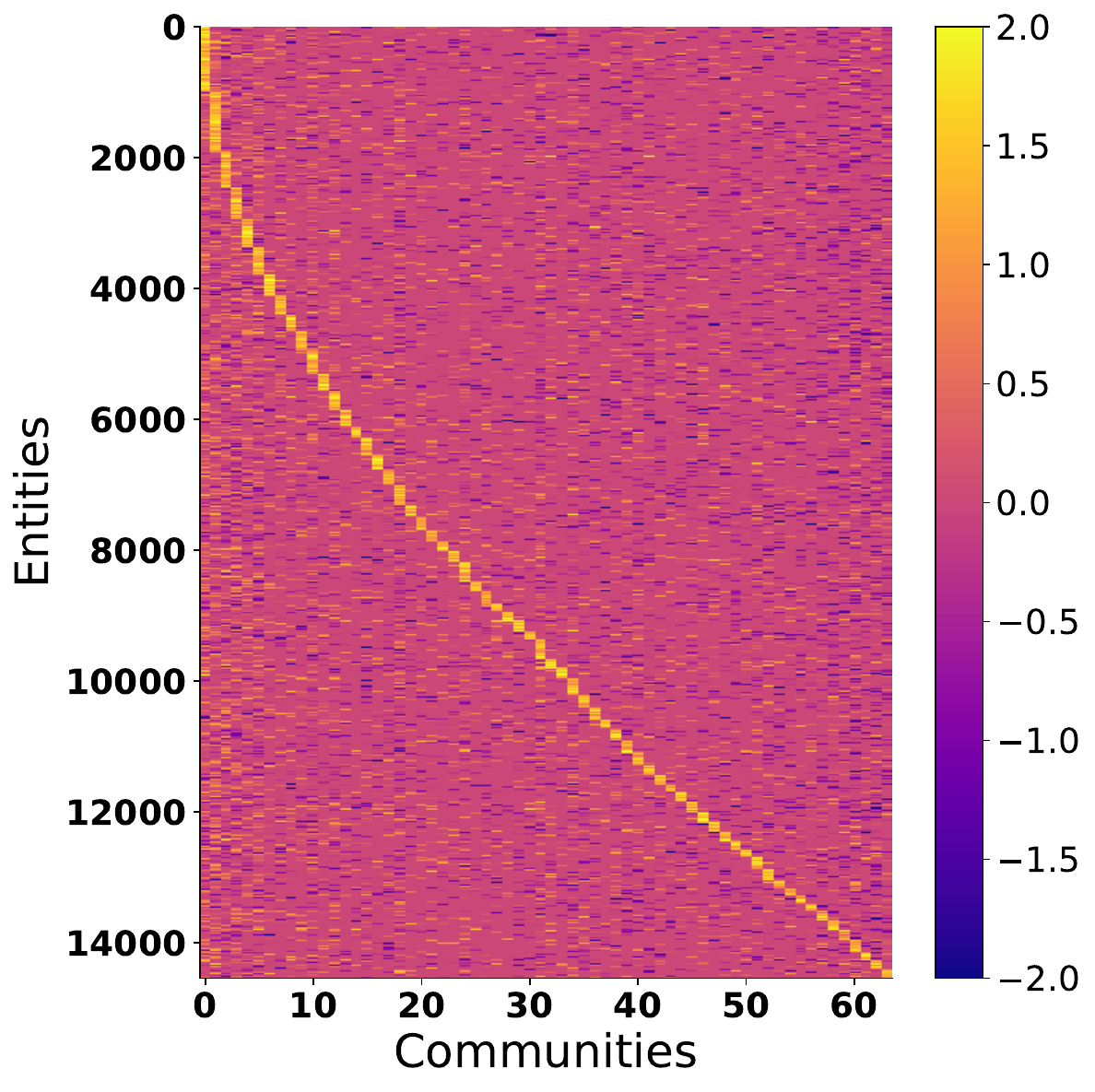}
        \label{fig:analysis_Fmatrix_fb}
    }
  \caption{The latent structure $F_\text{ans}$ learned from the WN18RR and FB15k-237 datasets. The columns of $F_\text{ans}$, representing communities, are sorted such that communities with higher summed strengths are assigned lower indices in the matrix.}
  \label{fig:analysis_Fmatrix}
\end{figure}

\textbf{Visualization of the latent structure}. For demonstration purposes, we use stick-breaking prior settings of $\alpha_\text{qry}=100$, $\alpha_\text{ans}=50$, and a truncation level of $K=64$ to generate the sparse latent feature matrix $F_\text{ans}$. In Figure \ref{fig:analysis_Fmatrix}, each matrix shows how entities are grouped into communities, where larger absolute values suggest stronger confidence in whether a node belongs to a specific community.

\newcommand{\cluster}[1]{\colorbox{cluster}{Cluster}: {#1}}
\newcommand{\entity}[1]{\newline{\colorbox{entity}{#1}}}

\begin{table}[h]
    \centering
    \vspace*{-4mm}
    \caption{Uncovered communities from the FB15k-237 dataset along with entity descriptions. \colorbox{cluster}{Community} and \colorbox{entity}{entity} names are highlighted in different colors, with entities in each community sorted in descending order by strength.}
    \noindent\fbox{
    \begin{minipage}{.75\linewidth}{\footnotesize
        \cluster{County}
        \entity{County Wexford}: County Wexford is a county in Ireland...
        \entity{Marion County}: Marion County is a county located in the U.S. state of Indiana...
        \entity{County Tyrone}: County Tyrone is one of the six counties of Northern Ireland...
        \\ \\
        \cluster{Music}
        \entity{PJ Harvey}: Polly Jean Harvey MBE is an English musician...
        \entity{Little Richard}: ...an American recording artist, songwriter, and musician...
        \entity{Italo disco}: Italo disco is a genre of music which originated in Italy...
        \entity{Talent manager-GB}: A talent manager, also known as band manager...
    }
    \end{minipage}
    }
  \label{tab:textual_communities_wn}
  \vspace*{-2mm}
\end{table}

\section{Related work}
\textbf{KGC with large language models (LLMs)}. Recent advancements in text-based KGC leverage the extensive pre-trained knowledge and contextual understanding of LLMs to bridge the gap between structured and unstructured knowledge. Techniques in this domain often employ diverse prompt designs to enable LLMs to perform direct reasoning for KGC \citep{yao2023exploring, wei2024kicgpt} or to refine textual information in datasets, enhancing their accuracy and richness \citep{li2023kermit, yang2024enhancing}. However, while these methods are training-free and inherently interpretable, they face challenges such as hallucinations and reliance on few-shot demonstrations, which are difficult to implement in sparsely connected KGs like WN18RR. Alternatively, some approaches fine-tune LLMs on KGC tasks using strategies like prefix-tuning \citep{chen2023dipping, zhang2024making} or adapter-tuning. While these methods capitalize on the reasoning capabilities of LLMs, they often lack interpretability, struggle to generalize across datasets, and continue to face challenges in achieving strong performance. In contrast, our model excels on relatively sparse KGs with distinct clustering patterns, leveraging text not only to improve KGC interpretability but also to provide meaningful clustering information about the KG itself. Additionally, while LLMs provide external knowledge to enhance KGC, our approach focuses on directly extracting and utilizing the intrinsic information within KGs to strengthen representation learning. This makes our method particularly effective in scenarios where LLMs cannot reliably provide external knowledge, such as in domain-specific datasets.

Our work also relates closely to \textbf{Variational AutoEncoders (VAEs)} \citep{kingma2013auto}, a foundational class of generative models that employs an encoder to map input data to a latent space, typically assuming a Gaussian prior, and a decoder to reconstruct the data from this latent representation. To facilitate gradient-based optimization during training, the reparameterization trick is used, re-expressing the sampling of latent variables as deterministic functions of noise variables, thereby enabling backpropagation through stochastic nodes. While this trick is straightforward for "location-scale" distributions like the Gaussian, extending it to other distributions such as Bernoulli \citep{jang2016categorical, maddison2016concrete} and Beta distributions \citep{nalisnick2016stick} requires more sophisticated techniques. Reparameterization for these distributions often involves implicit differentiation methods to compute gradients when explicit reparameterization is infeasible \citep{figurnov2018implicit}. A persistent challenge in training VAEs is posterior collapse, where the encoder’s output becomes similar to the prior, causing the model to ignore the latent variables \citep{bowman2015generating}. This issue undermines the VAE’s ability to learn meaningful representations. Various strategies have been proposed to mitigate posterior collapse, including modifying the objective function with $\beta$\-VAE to balance reconstruction and regularization terms \citep{higgins2017beta}, employing annealing schedules for the KL divergence term \citep{bowman2015generating}, and designing more expressive posterior distributions to better capture the underlying data structure \citep{rezende2015variational}.

% \newpage
% \input{body/checklist}

\end{document}